\documentclass[runningheads]{llncs}

\usepackage{eccv}

\usepackage{eccvabbrv}

\definecolor{tAccentA}{RGB}{255,140,0}
\definecolor{tAccentB}{RGB}{220,20,60}  
\definecolor{myyellow}{RGB}{230,180,0} 
\definecolor{mygreen}{RGB}{51,204,153} 
\definecolor{heat1}{HTML}{62BC7A}
\definecolor{heat2}{HTML}{A1CF7E}
\definecolor{heat3}{HTML}{DFE282}
\definecolor{heat4}{HTML}{FCD07E}
\definecolor{heat5}{HTML}{F99C74}
\definecolor{heat6}{HTML}{F7686A}

\usepackage{textcomp}
\usepackage{pifont}
\newcommand{\TitleIcon}{\raisebox{-0.15em}{\includegraphics[height=2.0em]{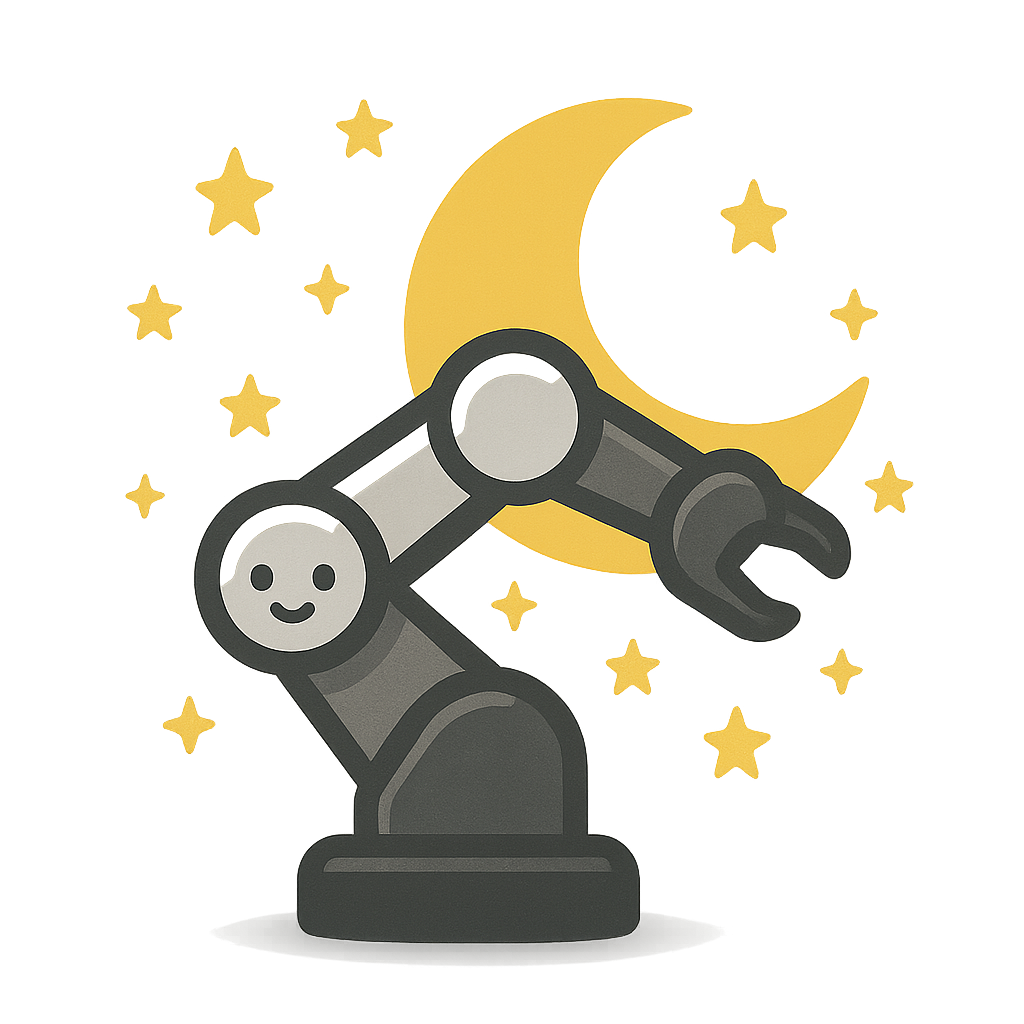}}}

\usepackage{booktabs}
\usepackage{multirow}
\usepackage[table]{xcolor}
\usepackage{siunitx}
\usepackage{makecell}
\usepackage{graphicx}       % for \resizebox
\usepackage{colortbl}       % \cellcolor/\columncolor
\usepackage{array}          
\usepackage{amsmath}
\usepackage{soul}

\definecolor{tblhead}{gray}{0.96}
\definecolor{oursrow}{gray}{0.94}
\definecolor{tblhead}{gray}{0.96}
\newcommand{\best}[1]{\textbf{#1}}          % best
\newcommand{\second}[1]{\underline{#1}}     % second
\definecolor{headergray}{gray}{0.95}

\DeclareMathOperator*{\argmax}{argmax}

\usepackage[accsupp]{axessibility}  %

\usepackage{hyperref}

\usepackage{orcidlink}

\makeatletter
\renewcommand*{\@fnsymbol}[1]{\ensuremath{\ifcase#1\or *\or \dagger\or \ddagger\or
    \mathsection\or \mathparagraph\or \|\or **\or \dagger\dagger
    \or \ddagger\ddagger \else\@ctrerr\fi}}
\makeatother

\begin{document}

\title{%
\texorpdfstring
    {
    \TitleIcon\
    \textcolor{tAccentA}{E\kern0.07em-}\kern-0.11em\textcolor{tAccentB}{VLA}:\
    \textcolor{tAccentA}{E}vent\kern0.02em-\kern-0.06emaugmented 
    \textcolor{tAccentB}{V}ision\kern0.02em-\kern-0.02em\textcolor{tAccentB}{L}anguage\kern0.06em-\kern-0.06em\textcolor{tAccentB}{A}ction Model for\\
    Dark and Blurred Scenes
    }
    {E-VLA:\ Event-Augmented Vision-Language-Action Model for Dark and Blurred Scenes}
}

\titlerunning{E-VLA}

\author{
Jiajun Zhai\inst{1,}\thanks{Equal contribution.}
\and 
Hao Shi\inst{2,1,*}
\and 
Shangwei Guo\inst{1}
\and 
Kailun Yang\inst{3}
\and 
Kaiwei Wang\inst{1,}\thanks{Correspondence.}
}

\authorrunning{J.~Zhai~\textit{et al.}}
\institute{$^1$Zhejiang University, $^2$Ant Group, $^3$Hunan University}

\maketitle

\begin{abstract}

Robotic Vision-Language-Action (VLA) models generalize well for open-ended manipulation, but their perception is fragile under sensing-stage degradations such as extreme low light, motion blur, and black clipping. We present E-VLA, an event-augmented VLA framework that improves manipulation robustness when conventional frame-based vision becomes unreliable. Instead of reconstructing images from events, E-VLA directly leverages motion and structural cues in event streams to preserve semantic perception and perception-action consistency under adverse conditions. We build an open-source teleoperation platform with a DAVIS346 event camera and collect a real-world synchronized RGB-event-action manipulation dataset across diverse tasks and illuminations. We also propose lightweight, pretrained-compatible event integration strategies and study event windowing for stable deployment. Experiments show that even a simple parameter-free fusion, \ie, overlaying accumulated event maps onto RGB images, could substantially improve robustness in dark and heavy-blur scenes: on Pick-Place at 20 lux, success increases from 0\% (image-only) to 60\% with overlay fusion and to 90\% with our event adapter; under severe motion blur (1000 ms-exposure proxy), Pick-Place improves from 0\% to 20-25\%, and Sorting from 5\% to 32.5\%. Overall, E-VLA provides systematic evidence that event-driven perception can be effectively integrated into VLA models, pointing toward robust embodied intelligence beyond conventional frame-based imaging. Code and dataset will be available at \href{https://github.com/JJayzee/E-VLA}{E-VLA}.


\end{abstract}    
\section{Introduction}
\label{sec:intro}

\begin{figure}[t]
    \centering
    \includegraphics[width=\linewidth]{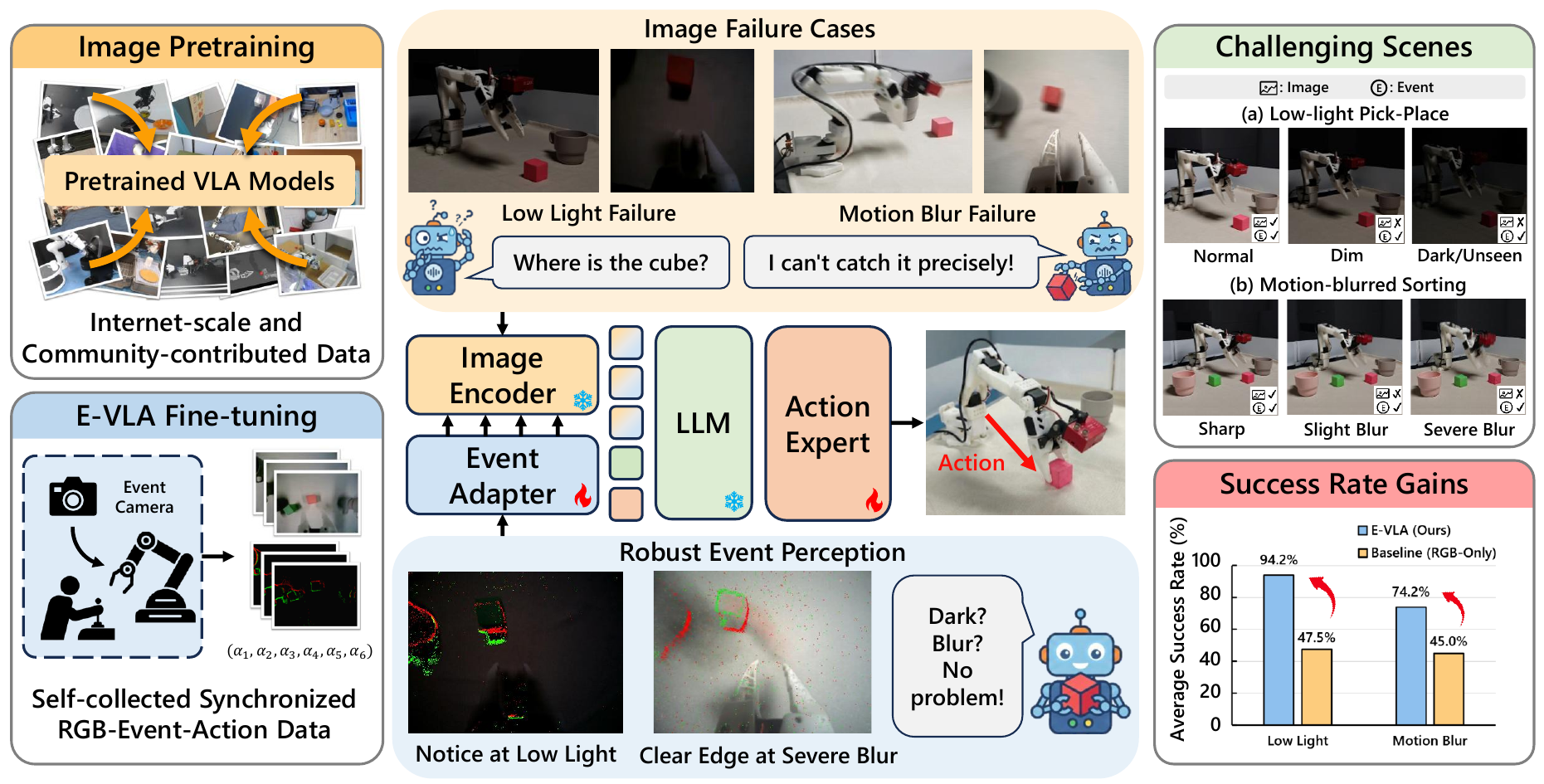}
    \vskip -2ex
    \caption{
    Middle: Image-based VLA models degrade under low-light and motion-blurred conditions, leading to failures in object detection and imprecise manipulation. We propose E-VLA, which fuses stable event visual cues with image features within the VLA pipeline, preserving reliable performance under adverse conditions.
    Left: Existing VLA models (\textit{e.g.}, SmolVLA~\cite{shukor2025smolvla}) are pretrained on large-scale text-image datasets and manipulation videos. 
    We build a teleoperation system equipped with an event camera and collect synchronized RGB-event-action data for E-VLA training.
    Right: We evaluate our method across gradient low-light and motion-blurred scenes. 
    E-VLA consistently achieves higher success rates than RGB baselines.}
    \label{fig:teaser}
    \vskip-4ex
\end{figure}

Vision-Language-Action (VLA) models have recently emerged as a promising paradigm for robotic manipulation, enabling systems to map visual observations and language instructions directly to action sequences. Representative systems such as RT-1~\cite{brohan2022rt1}, RT-2~\cite{zitkovich2023rt2}, the $\pi$ family~\cite{black2024pi_0,intelligence2025pi_0.5,intelligence2025pi_0.6}, OpenVLA~\cite{kim2024openvla}, RDT-1B~\cite{liu2024rdt}, and Gemini Robotics~\cite{team2025gemini} demonstrate a common trend: leveraging large-scale robot interaction data and pretrained vision-language models to learn versatile, language-conditioned manipulation policies. 
Most existing VLA approaches are built upon frame-based RGB perception, where pretrained visual encoders are coupled with policy learning modules to jointly model perception, reasoning, and control, and have shown strong performance on open-ended manipulation tasks in well-lit laboratory settings~\cite{ma2024survey,zhong2025survey,li2025survey,deng2025SurveyRL}.

However, real-world deployment exposes a critical weakness of current VLA systems: perceptual robustness under sensing-stage degradations. Recent surveys and benchmarks highlight illumination variation and visual domain shift as major factors limiting stable VLA performance~\cite{ma2024survey,li2025survey,libero-plus2025}. In particular, low-light conditions substantially reduce signal quality, while increasing exposure time to recover brightness inevitably introduces motion blur and additional latency during fast manipulation. In extreme cases, severe under-exposure can lead to black clipping, where frame images become nearly unusable. Importantly, these failures originate from the limitations of frame-based imaging itself. 
Although image enhancement and data augmentation can improve visual appearance, they work on already degraded measurements and cannot recover information lost at capture time.
Event cameras, which asynchronously capture brightness changes with high temporal resolution and wide dynamic range~\cite{delbruck2008framefree,gallego2020eventsurvey}, offer a compelling alternative for robust robotic perception under low light and fast motion. 
Yet integrating event sensing into VLA is nontrivial. 
First, event streams are sparse and statistically different from dense RGB images, making them poorly matched to image-domain pretraining that modern VLA models rely on. Second, in manipulation settings, event triggering is tightly coupled with arm motion and control timing, resulting in highly non-stationary temporal distributions. These challenges make it unclear how to incorporate event signals into pretrained VLA architectures while preserving stable perception--action coupling.

To address this gap, we introduce \textbf{E-VLA}, an event-augmented VLA framework for robust manipulation under visually degraded conditions. We build an open-source teleoperation platform equipped with a DAVIS346 event camera and collect a real-world synchronized RGB-event-action dataset across multiple manipulation tasks and illumination conditions, enabling systematic study of event-augmented VLA learning. 
On top of a SmolVLA-style backbone, we design lightweight and pretrained-compatible event integration strategies, including a parameter-free event overlay mechanism and a hierarchical event adapter, and investigate event windowing/accumulation schemes to manipulation dynamics.

Extensive experiments show that E-VLA substantially improves robustness in low-light and blur-prone scenarios while preserving performance in normal lighting. Notably, under complete black clipping (an extreme low-light failure mode), E-VLA maintains task success rates above 80\%, compared with 0\% for the image-only baseline. We further show that event-count-based accumulation yields more stable perception--action coupling than time-based windowing, and that event augmentation improves generalization to unseen degraded conditions without requiring additional low-light training data.

To summarize, our contributions are as follows:
\begin{itemize}
    \item We present \textbf{E-VLA}, the first Vision-Language-Action framework that integrates event-based visual sensing for robust robotic manipulation under low-light and motion-blur degradations.
    \item We build an {open-source event-augmented teleoperation platform} and collect a {real-world synchronized RGB-event-action manipulation dataset} across multiple tasks and illumination conditions, enabling systematic validating and training of event-enhanced VLA models.
    \item We propose {lightweight, pretrained-compatible event integration strategies}, including a parameter-free overlay fusion and a hierarchical event adapter, together with a manipulation-aware event windowing design.
    \item We conduct extensive experiments and ablations, providing the first systematic empirical evidence and practical design insights for integrating event-driven perception into scalable VLA learning.
\end{itemize}
\section{Related Work}
\label{sec:related_works}

\subsection{Vision-Language-Action Systems}
Vision-Language-Action (VLA) models learn large-scale policies that directly map visual observations and language instructions to robot actions~\cite{ma2024survey,zhong2025survey,li2025survey}. Representative systems, including the RT series~\cite{brohan2022rt1,zitkovich2023rt2}, the $\pi$ family~\cite{black2024pi_0,intelligence2025pi_0.5,intelligence2025pi_0.6}, OpenVLA~\cite{kim2024openvla}, RDT-1B~\cite{liu2024rdt}, and Gemini Robotics~\cite{team2025gemini}, demonstrate that scaling multimodal transformers with large robot datasets enables generalizable language-conditioned manipulation across diverse tasks. Recent efforts further explore enhanced spatial reasoning via 3D representations~\cite{li2025pointvla,singh2025ogvla,sun2025geovla,deng2025stereovla}, tactile integration for contact modeling~\cite{huang2025tactile,bi2025vlatouch,cheng2025omnivtla}, reinforcement learning refinement~\cite{guo2025irevla,lu2025vlarl,chen2025tgrpo}, and efficient deployment strategies~\cite{wen2025tinyvla,shukor2025smolvla,wang2025vlaadapter}. 
Despite architectural diversity, existing VLA systems predominantly rely on frame-based RGB cameras as their primary perceptual interface. Frame-based imaging integrates photons over an exposure interval, which inherently limits signal-to-noise ratio in low illumination and induces motion blur during fast movements. Such degradations occur at the sensing stage and propagate through pretrained visual encoders, ultimately affecting policy stability. While geometric or tactile extensions improve spatial reasoning and interaction modeling, they do not fundamentally address the imaging-level limitations of frame-based perception. As a result, the perceptual robustness of VLA models under extreme illumination and high-speed motion remains insufficiently explored.

\subsection{Event-Based Perception in Robotics}
Event cameras asynchronously record per-pixel brightness changes with microsecond latency and high dynamic range~\cite{delbruck2008framefree,gallego2020eventsurvey}. By avoiding exposure-based temporal integration, event sensing naturally mitigates motion blur and maintains signal fidelity under extreme lighting conditions. These properties have motivated extensive research in event-based perception, including image reconstruction~\cite{rebecq2019e2vid,cadena2021spade-e2vid}, low-light enhancement~\cite{liang2024evlight,chen2025evlightpp,sun2025lowlight}, motion deblurring~\cite{pan2019edi,shang2021d2net,sun2022efnet,lu2025uniinr}, segmentation~\cite{alonso2019ev_segnet,li2024eventseg}, detection~\cite{tomy2022fusing_adverse,cao2024chasing}, optical flow~\cite{zihao2018unsupervised,gehrig2021eraft,ye2023towards_event}, and tracking~\cite{wang2024towards_robust_tracking,liu2025tracking}. In robotics, event sensing has been applied to visual servoing and grasping~\cite{li2020eventgrasp,muthusamy2021neurograsp,huang2022visiongrasp}, navigation~\cite{neuromorphicnav2025,bugueno2025humanrobotnav}, slip detection~\cite{reinold2025simuslip,guo2024force}, and visual–tactile perception~\cite{taunyazov2020eventtact,funk2024evetac}. 
However, most existing event-based robotic systems are designed for specific tasks and rely on modular pipelines with task-dependent representations and predefined control strategies. In contrast, modern VLA models aim to learn general-purpose, language-conditioned manipulation policies through large-scale multimodal pretraining. Despite the complementary sensing advantages of event cameras, their integration into scalable VLA frameworks remains largely unexplored. Bridging event-driven perception with pretrained VLA architectures therefore presents a promising direction for enhancing embodied robustness without sacrificing policy generality.
\section{Methods}
\label{sec:methods}

\subsection{Overview}
We build upon a SmolVLA-style~\cite{shukor2025smolvla} Vision-Language-Action framework and investigate how event-based visual sensing can be integrated in a lightweight and pretrained-compatible manner. Event streams captured by a DAVIS camera are spatially and temporally aligned with RGB frames and converted into frame-like representations through carefully designed windowing strategies, ensuring compatibility with standard visual encoders.

\begin{figure}[t]
    \centering
    \includegraphics[width=\linewidth]{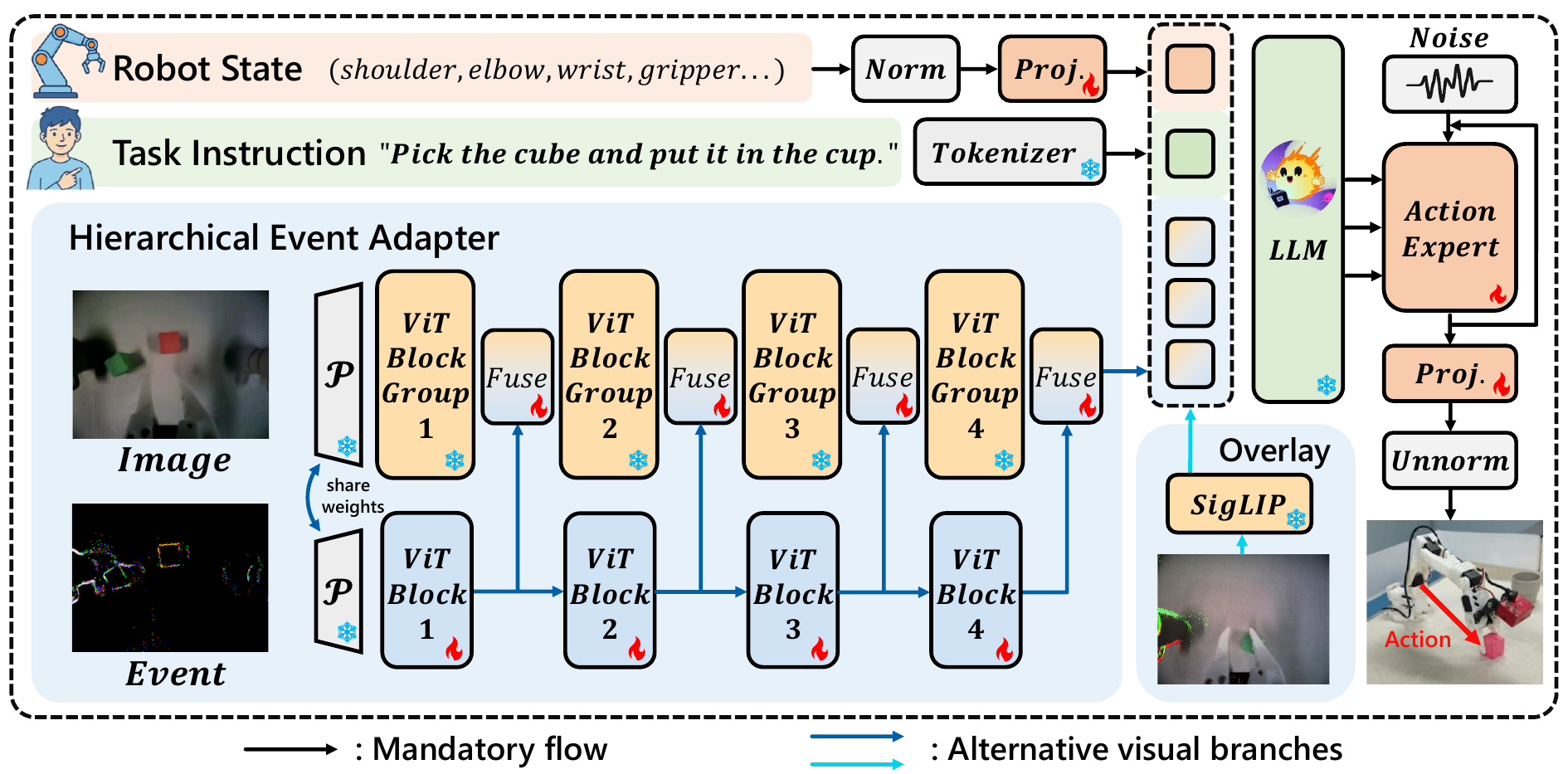}
    \vskip-1ex
    \caption{
    An overview of our proposed E-VLA framework. Our architecture integrates event-based visual sensing with RGB frames and proprioceptive robot states to generate control sequences. We investigate two fusion strategies: (1) a Hierarchical Event Adapter that injects event features into intermediate layers of a frozen ViT encoder through trainable fusion modules, and (2) an Overlay strategy that directly combines events with RGB images prior to encoding via SigLIP. The resulting fusion visual tokens are concatenated with language tokens and state tokens and then processed by a frozen LLM backbone, which conditions an Action Expert to produce normalized robot actions. Snowflakes and flames denote frozen and trainable parameters, respectively.
}
    \label{fig:pipeline}
    \vskip-3ex
\end{figure}

\begin{figure}[t]
    \centering
    \includegraphics[width=\linewidth]{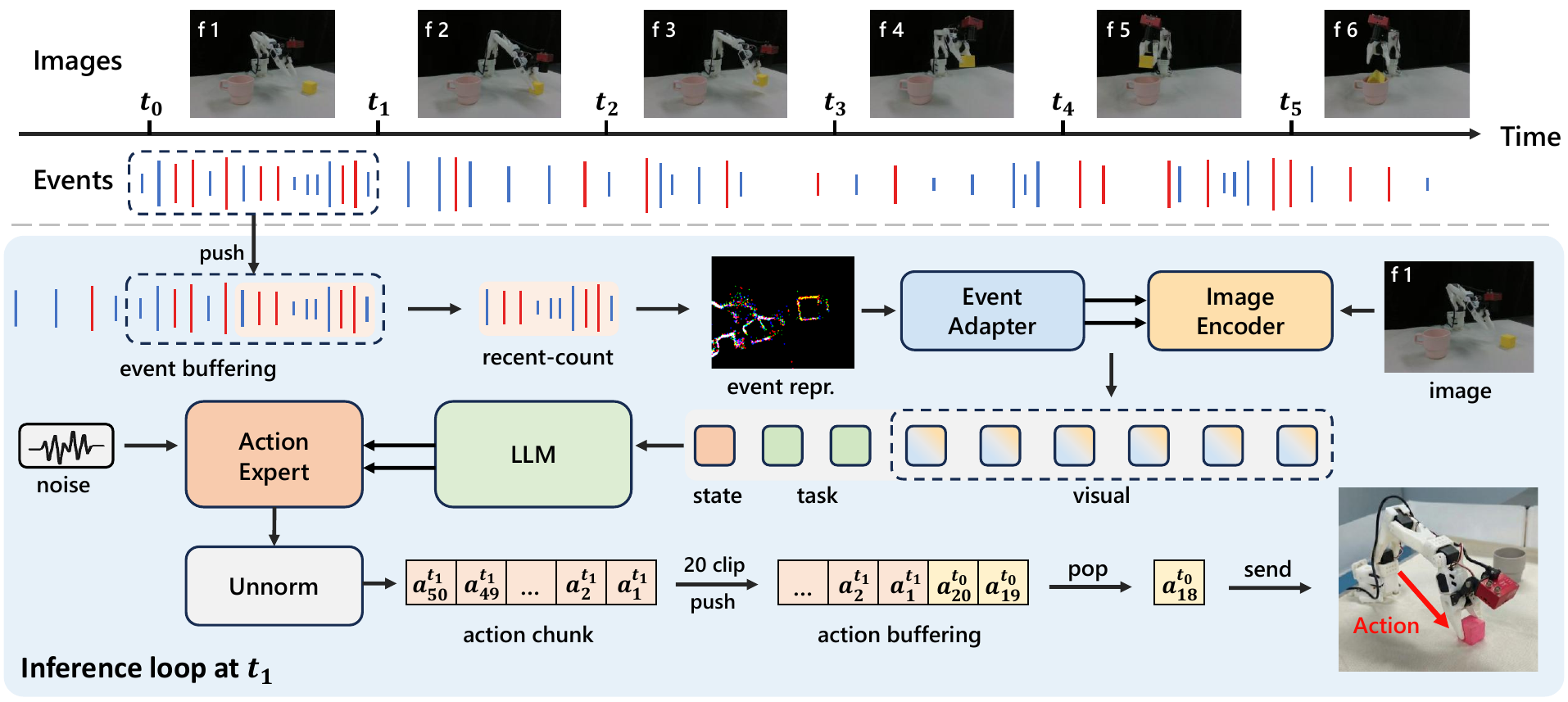}
    \vskip-1ex
    \caption{
    A system schematic of our E-VLA pipeline. Incoming events are first buffered and aggregated via a recent-count representation, then converted into an event representation and aligned through an event adapter with visual features extracted by an image encoder. The fused multimodal visual tokens are then fed into LLM, which interacts with an action expert to generate chunked action outputs. The predicted actions are stored in an action buffer before being executed in a streaming manner.
}
    \label{fig:schematic}
    \vskip-3ex
\end{figure}

To incorporate event information without disrupting the pretrained token distribution, we investigate two alternative fusion designs that operate at different stages of the visual pipeline. The first design adopts a pre-encoding overlay strategy, directly combining accumulated event frames with RGB images before visual encoding, introducing no additional parameters. Instead of overlaying, the second design employs a lightweight hierarchical event adapter to inject event features into intermediate layers of the visual encoder for fine-grained cross-modal interaction. In each case, the resulting visual tokens are processed by the VLM backbone, and an action expert generates control sequences conditioned on the enriched perceptual representations. An overview of our E-VLA framework is depicted in Fig.~\ref{fig:pipeline} and a system schematic is shown in Fig.~\ref{fig:schematic}. The following sections detail the baseline architecture, event windowing and representation design, and the proposed fusion strategies.

\subsection{SmolVLA Baseline}

We adopt SmolVLA~\cite{shukor2025smolvla} as the image-based policy baseline to have a good balance between performance and efficiency. SmolVLA is an open-source lightweight vision-language-action model trained on community-contributed datasets with about 0.5B parameters. 
It offers strong reproducibility compared to larger or closed-source VLA models, while maintaining competitive performance in real-world robot deployment.
SmolVLA consists of a compact vision-language backbone (SmolVLM~\cite{marafioti2025smolvlm}) and a lightweight transformer-based action expert. Using SigLIP as its vision encoder, SmolVLM is pretrained on large-scale text-image and video data, encoding multi-view RGB observations, language instructions, and robot states into a unified token representation. The action expert then generates action chunks conditioned on SmolVLM features through the interaction design of interleaved self-attention and cross-attention. SmolVLA supports synchronous and asynchronous inference modes. In synchronous inference, a new policy inference process is only triggered when the current action queue is empty, which has a low perception delay while introducing execution pauses due to inference latency. In contrast, asynchronous mode performs inference in parallel with action execution, eliminating explicit waiting between action chunks while prolonging its perceptual delay. Overall, this established image baseline provides a solid foundation for our event-augmented VLA approach.

\subsection{Event Windowing and Representation}
\label{subsec:win_and_repr}

Event cameras produce asynchronous and spatially sparse event streams, which record pixel-level brightness changes. The output is formulated as tuples:
\begin{equation}
    \label{eq:event_tuple}
    \mathcal{E}=\{e_i\}_{i=1}^M, \quad e_i=(x_i, y_i, t_i, p_i).
\end{equation}
 
where $x_i$ and $y_i$ denote the pixel coordinates of the event, $t_i$ denotes its timestamp, and $p_i\in{+1,-1}$ is the polarity of the brightness change. The output of this format cannot be directly consumed by standard vision-language models. Prior event-based vision work has explored windowing and representation strategies to convert raw event streams into structured forms, including fixed-duration windows and fixed-count windows for sampling, as well as event frames, voxel grids~\cite{zihao2018unsupervised}, time surfaces~\cite{lagorce2016timesurface}, and other specific methods~\cite{alonso2019ev_segnet,sun2024motion_representation} for representation.
Fixed-duration windowing is widely used in event-based methods~\cite{cadena2021spade-e2vid,li2024eventseg,tomy2022fusing_adverse,gehrig2021eraft,liu2025tracking} to maintain a frame rate or to synchronize with image exposure time.

However, in the VLA setting, the temporal distribution of event streams becomes highly unstable due to the tight coupling between perception and action. The wrist-mounted camera undergoes continuous ego-motion induced by robot actions, and the event rate varies substantially with the frequently changing motion speed across task phases. Under such conditions, event windowing becomes non-trivial: fixed-duration windows may produce inconsistent event responses, leading to perception failure at low speeds and motion blur at high speeds. Since the suitability of fixed-duration versus fixed-count windowing in VLA scenarios remains unclear, we empirically examine these strategies and adopt a \emph{recent-count} scheme that retains the most recent $N$ events, which can be formulated as Eq.~\ref{eq:recent_count_window}, where $t_e$ denotes the exposure end time for the current image $I\in\mathbb{R}^{H\times W\times 3}$, $\mathcal{W}_{I}$ denotes the event window corresponding to $I$. 
\begin{equation}
    \label{eq:recent_count_window}
    \mathcal{E}_{t_e}=\{e_i \mid t_i\le t_e \}, \quad 
    \mathcal{W}_I=\{e_k\}_{k=|\mathcal{E}_{t_e}|-N+1}^{|\mathcal{E}_{t_e}|}.
\end{equation}
This model demonstrates better event stability in practice. The DAVIS sensor provides events and RGB images with microsecond-resolution timestamps, enabling precise temporal alignment for implementing this windowing operation.

Compared to large-scale image-text and video datasets, event data remain limited in both scale and diversity, making large-scale pretraining or contrastive learning directly on event modalities impractical. To effectively leverage the knowledge embedded in pretrained VLMs, we adopt an image-like accumulated event representation that closely matches the input distribution of the SigLIP image encoder used in SmolVLA. Within each window, events are accumulated into a dense gray frame $\tilde{E}\in \mathbb{R}^{H\times W}$ via a polarity-agnostic way: the total amount of events is added up at each pixel regardless of polarity. This representation emphasizes event density and the absolute position while discarding the direction of intensity change, resulting in a structurally stable input format. We use a colored DAVIS346 camera equipped with a Bayer filter, enabling the capture of RGB event streams. So the accumulated event frames are first normalized to $[0,1]$ and then demosaiced to obtain three-channel representations $E\in \mathbb{R}^{H\times W\times 3}$. The whole process can be formulated as Eq.~\ref{eq:event_frame}, where $\delta(x,y)$ denotes the 2D Kronecker delta function, $\text{Norm}(\cdot)$ denotes the min-max normalization, and $\mathcal{D}(\cdot)$ denotes the demosaicing operation. 
\begin{equation}
    \label{eq:event_frame}
    \tilde{E}(x,y) = \sum_{e_i\in\mathcal{W}_I}\delta(x-x_i, y-y_i), \quad E=\mathcal{D}(\text{Norm}(\tilde{E})).
\end{equation}

\subsection{Event Fusion Strategy}

In the VLA architecture, visual observations are encoded into a sequence of tokens by a ViT-style encoder, then concatenated with language instruction tokens and robot state tokens before being fed into the LLM. Under this formulation, naively introducing additional event tokens would increase the token length and computational cost, and potentially disrupt the pretrained token distribution of the visual encoder. To preserve the original token length and leverage pretrained visual-language knowledge, we fuse event information into the existing visual tokens, rather than adding new event tokens.

We explore two event fusion strategies that operate at different stages of the visual processing pipeline: an overlay-based fusion that combines events and images before visual encoding, and a lightweight hierarchical event adapter that integrates event features within the visual encoder. 

\noindent \textbf{Overlay-based fusion.} 
Given the spatial alignment between frame images and events provided by the DAVIS camera, a simple pre-encoding fusion can be implemented by directly overlaying the windowed events onto RGB images. Let $\mathcal{E}_{(x,y)}$ denotes the events at position $(x,y)$ within the window: 
\begin{equation}
    \label{eq:ev_in_window}
    \mathcal{E}_{(x,y)}=\{e_i \mid e_i \in \mathcal{W}_I, \ (x_i,y_i)=(x,y) \}.
\end{equation}
The overlay process can be formulated as Eq.~\ref{eq:overlay}, where $I^{\text{o}}$ donates the overlay image, $\boldsymbol{c}(p_j)$ donates the polarity-to-color projection. 
\begin{equation}
    \label{eq:overlay}
      I^{\text{o}}(x,y)=
      \begin{cases}
        I(x,y), & |\mathcal{E}_{(x,y)}|=0, \\
        \boldsymbol{c}(p_j), & j=\argmax_i t_i, \ e_i\in\mathcal{E}_{(x,y)}.
      \end{cases}
\end{equation}
This approach directly injects events into the visual input before ViT encoding, enriching the visual information without adding parameters, making it a plug-and-play and computationally efficient implementation.

\noindent \textbf{Hierarchical event adapter.} 
To enable progressive and fine-grained fusion of event and image information, we further propose a lightweight hierarchical event adapter that integrates event features into the image encoder during visual encoding. Specifically, accumulated event frames $E$ are processed by a compact ViT-style event adapter which consists of a $16\times16$ weight-sharing patch embedding layer (sharing weights for closer feature distribution) and four stacked Transformer blocks. 
The outputs of these blocks are progressively fused with the corresponding intermediate features of the original visual encoder. For SigLIP in SmolVLA, we choose $3$, $6$, $9$, and $12$ as fusion layers.
Let $\mathcal{G}_l(\cdot)$ denotes the $l$-th layer of the event adapter, $\mathcal{F}_l(\cdot)$ denote the $l$-th group of the SigLIP encoder layers, $E^{(l)}$ and $F^{(l)}$ denote their output features, respectively,  and $\mathcal{P}(\cdot)$ denotes the weight-sharing patch embedding. Initially, we have $E^{(0)}=\mathcal{P}(E)$, $F^{(0)}=\mathcal{P}(I)$. Then the overall hierarchical fusion process can be formulated as Eq.~\ref{eq:adapter}, where $\text{Fuse}(\cdot)$ is implemented as $\text{MLP}(\text{Concat}(\cdot))$.
\begin{equation}
    \label{eq:adapter}
    E^{(l+1)}=\mathcal{G}_{l+1}(E^{(l)}), \ \
    F^{(l+1)}=
        \mathcal{F}_{l+1}(\text{Fuse}(F^{(l)},E^{(l)}))).
\end{equation}
 The introduced event adapter and fusion module adopt smaller hidden dimensions and fewer layers, which only introduce 13M additional parameters, less than 3\% of the total model, enabling a fine-grained and multi-level event-image fusion while preserving computational efficiency.
\section{Dataset}
\label{sec:dataset}

\begin{figure}[t]
    \centering
    \includegraphics[width=\linewidth]{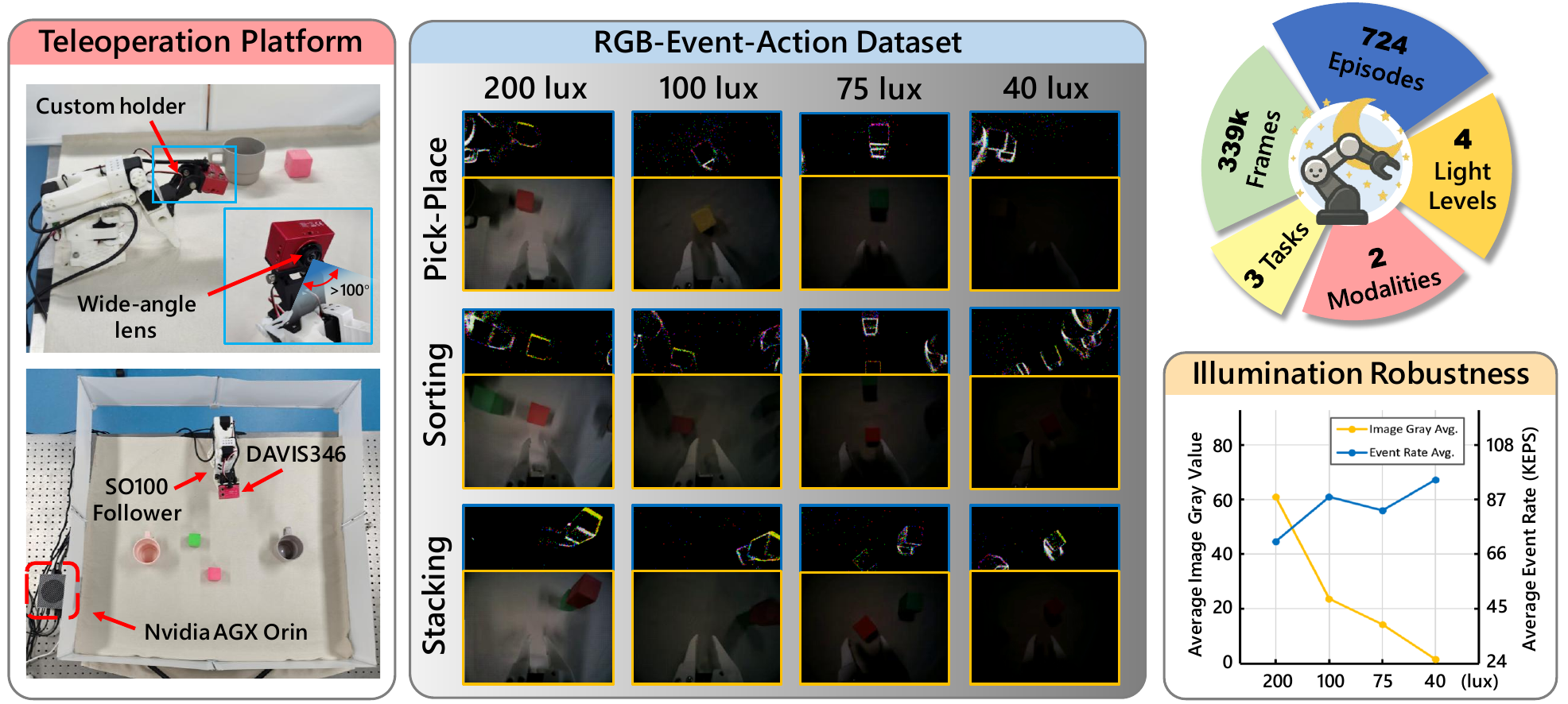}
    \caption{
    Middle: The visualization of the proposed dataset. Events are represented as colored frames following Sec.~\ref{subsec:win_and_repr}. Left: Side and top views of our teleoperation platform based on the LeRobot SO100 manipulator~\cite{cadene2024lerobot} and the DAVIS346 event camera. Right: Above are the statistics of our dataset. The line chart below shows that even when the image signal rapidly decays with decreasing illumination, the event modality can still maintain a stable event rate ($\sim 87k$ events per second or KEPS).
}
    \label{fig:platform_dataset}
    \vskip-3.5ex
\end{figure}

\subsection{Event-Augmented Teleoperation Platform}

To support event-augmented VLA learning and real-world evaluation under visually degraded conditions, we build a unified teleoperation platform on the open-source SO100 6-DoF robotic arm~\cite{cadene2024lerobot}, using structurally identical leader--follower arms for demonstration collection and closed-loop testing. The follower arm is equipped with an angle-driven gripper as the end-effector, and a DAVIS346 event camera is rigidly mounted on the wrist via a custom holder, providing pixel-aligned asynchronous events and RGB frames at $346\times260$ resolution with microsecond-level timestamps. Instead of adopting a multi-view setup, we deliberately use a monocular wrist-view camera to preserve system compactness and deployment practicality, while naturally coupling arm motion with event generation, providing an appropriate setting for integrating event cues to explore perception--action robustness under visual degradation.
In practice, we find that a wide field of view is critical for wrist-view manipulation, as narrow-view lenses frequently miss task-relevant context; we therefore replace the original pinhole lens with a compact wide-angle lens (2.2\,mm focal length, F/1.8), achieving a field of view greater than $100^\circ$ with suitable depth of field for typical manipulation distances. The DAVIS camera operates with fixed hardware gain; all frames are recorded at 30\,FPS, and unless otherwise specified, we use a 10\,ms exposure time. To reflect realistic edge deployment constraints, both data collection and on-robot inference are conducted on an NVIDIA AGX Orin platform.

\subsection{RGB-Event-Action Dataset}

Existing event-based datasets~\cite{nah2017gopro,stoffregen2020hqf,gehrig2021dsec,sun2023highrev,li2024flowtrackingdataset,bao2024temporal} mainly focus on perception tasks (\textit{e.g.}, deblurring, optical flow, or reconstruction), while real-world event datasets for imitation learning and VLA-style manipulation remain extremely limited. 
To bridge this gap, we collect a new event-augmented manipulation dataset using the teleoperation system above; to the best of our knowledge, this is among the first real-world datasets designed specifically for studying synchronized \emph{RGB--event--action} learning in language-conditioned manipulation under challenging illumination. 
Following SmolVLA~\cite{shukor2025smolvla}, we include three representative tasks: \textbf{Pick-Place}, \textbf{Sorting}, and \textbf{Stacking}. Pick-Place requires grasping an object from a random location and placing it into a fixed cup; Sorting additionally requires color-based discrimination to place objects into designated cups; and Stacking requires precise placement of one cube on top of another under a constrained gripper-centric field of view. For each task, demonstrations are collected under one normal illumination condition (200\,lux) and three lower illumination levels (100, 75, and 40\,lux), with both fixed and randomized initial arm configurations to improve diversity and robustness. Due to the low hardware gain of the DAVIS camera, bright scenes under 200\,lux still produce slightly dark images with about $60$ of $255$ grayscale value as shown in Fig.~\ref{fig:platform_dataset}. All data are stored in the LeRobot format, where RGB frames and event streams are temporally aligned by camera timestamps and synchronized with robot states and actions. 
In summary, our dataset comprises $724$ episodes ($305$ under normal illumination and 419 under three progressively decreasing low-light levels), totaling 339310 frames. Fig.~\ref{fig:platform_dataset} provides an overview of the dataset composition and capture setup.

\section{Experiment}
\label{sec:Exp}

\subsection{Implementation Details}

We compare E-VLA against four representative alternatives: RetinexNet, Retinexformer, EvLight, and E2VID, using official pretrained weights and standard implementations. RetinexNet, Retinexformer, and EvLight are treated as low-light image enhancement baselines, where low-illumination images are enhanced offline for training and processed with the same enhancement pipeline at inference time; E2VID is used as an event-to-image reconstruction baseline, where event streams are first converted into color images and then introduced as additional visual inputs during both training and testing. For fair comparison, the VLM backbone is kept frozen in all settings, and imitation learning updates only the action expert and projection MLP layers. For our event-adapter variant, we also freeze the VLM backbone to preserve pretrained visual-language priors, and train the event branch with a two-stage strategy: we first optimize the newly introduced adapter from scratch, and then jointly fine-tune the adapter together with the action expert and projection layers. To avoid shortcut learning from RGB inputs and encourage effective event utilization, we apply a random dropout to the image branch during training. Unless otherwise specified, all methods are fine-tuned on the same multi-task, multi-illumination dataset for 20k iterations, and training is performed on a single NVIDIA A800 GPU.

\subsection{Evaluation Protocol}
We evaluate all methods on the same real-world teleoperation platform (SO100 + DAVIS346) over three manipulation tasks (Pick-Place, Sorting, Stacking) under multiple illumination conditions, with policy inference executed on an NVIDIA AGX Orin. For fair comparison, we sample task-specific test configurations once and keep them fixed: Pick-Place uses 10 test locations, and Sorting/Stacking uses 10 paired configurations with randomly swapped object placements. We adopt partial-credit scoring (Pick-Place: pick/place; Sorting: per-object pick/place; Stacking: pick/stack). We use synchronous inference with an action chunk size of 40 at 30\,Hz (predicting $\sim$1.33\,s actions per forward pass). Success is measured within 1 minute for Pick-Place/Stacking and 1.5 minutes for Sorting. Additional implementation details are provided in the supplementary material.

% Hao
\begin{table}
  \centering
  \setlength{\tabcolsep}{4pt}
  \renewcommand{\arraystretch}{1.15}
  \caption{Task success rates (\%). Higher is better. Results are reported under different illumination levels (lux) on the Pick-Place task.}
  \vskip-2ex
  \label{tab:1_main_results}
    
  % 1 列方法名 + 6 列数字（整数示例；若有一位小数把 table-format=3 改为 3.1）
  \resizebox{\textwidth}{!}{%
    \begin{tabular}{@{} l *{6}c S[table-format=2.1]}
      \toprule
      \rowcolor{tblhead}
      \multicolumn{1}{c}{\textbf{Method}} &
      \multicolumn{1}{c}{\textbf{75~lux}} &
      \multicolumn{1}{c}{\textbf{40~lux}} &
      \multicolumn{1}{c}{\textbf{35~lux}} &
      \multicolumn{1}{c}{\textbf{30~lux}} &
      \multicolumn{1}{c}{\textbf{25~lux}} &
      \multicolumn{1}{c}{\textbf{20~lux}} &
      \multicolumn{1}{c}{\textbf{Average}} \\
      \midrule
      \midrule
      \multicolumn{1}{l}{Image (10ms exp.)}                                 & \cellcolor{heat1}100 & \cellcolor{heat2}80 & \cellcolor{heat3}70 & \cellcolor{heat6}35 & \cellcolor{heat6}0 & \cellcolor{heat6}0 & 47.5 \\
      \multicolumn{1}{l}{Image+RetinexNet~\cite{wei2018retinexnet}}         & \cellcolor{heat1}100 & \cellcolor{heat1}100 & \cellcolor{heat2}85 & \cellcolor{heat2}80 & \cellcolor{heat6}25 & \cellcolor{heat6}10 & 66.7 \\
      \multicolumn{1}{l}{Image+Retinexformer~\cite{cai2023retinexformer}}   & \cellcolor{heat1}100 & \cellcolor{heat2}80 & \cellcolor{heat2}80 & \cellcolor{heat3}75 & \cellcolor{heat6}20 & \cellcolor{heat6}10 & 60.8 \\
      \addlinespace[2pt]
      \midrule
      \addlinespace[2pt]
      \multicolumn{1}{l}{Image+EvLight~\cite{liang2024evlight}}     & \cellcolor{heat1}100 & \cellcolor{heat1}95 & \cellcolor{heat1}95 & \cellcolor{heat3}75 & \cellcolor{heat6}45 & \cellcolor{heat6}10 & 70.0 \\
      \multicolumn{1}{l}{Image+E2VID~\cite{rebecq2019e2vid}}        & \cellcolor{heat2}80 & \cellcolor{heat4}60 & \cellcolor{heat5}55 & \cellcolor{heat6}10 & \cellcolor{heat6}5 & \cellcolor{heat6}5 & 35.8 \\
      \rowcolor{oursrow}
      \multicolumn{1}{l}{Ours overlay}                              & \cellcolor{heat1}100 & \cellcolor{heat1}100 & \cellcolor{heat2}85 & \cellcolor{heat3}75 & \cellcolor{heat4}65 & \cellcolor{heat4}60 & \second{80.8} \\
      \rowcolor{oursrow}
      \multicolumn{1}{l}{Ours event adapter}                        & \cellcolor{heat1}100 & \cellcolor{heat1}100 & \cellcolor{heat1}95 & \cellcolor{heat1}90 & \cellcolor{heat1}90 & \cellcolor{heat1}90 & \best{94.2} \\
      \bottomrule
    \end{tabular}%
  }% <-- \resizebox 结束
  \vspace{3pt}
    \begin{minipage}{\textwidth}
        \centering 
        \tiny \textit{Success rates (\%):} \hspace{0.25em}
        \colorbox{heat1}{\phantom{xx}} $\ge 90$ (\textbf{Best}) \hspace{0.25em}
        \colorbox{heat2}{\phantom{xx}} $80 \sim 90$ \hspace{0.25em}
        \colorbox{heat3}{\phantom{xx}} $70 \sim 80$ \hspace{0.25em}
        \colorbox{heat4}{\phantom{xx}} $60 \sim 70$ \hspace{0.25em}
        \colorbox{heat5}{\phantom{xx}} $50 \sim 60$ \hspace{0.25em}
        \colorbox{heat6}{\phantom{xx}} $\le 50$ (Bad)
    \end{minipage}
\vskip-3ex
\end{table}

\subsection{Low Illumination Performance}
We evaluate task success rates under progressively reduced ambient illumination for image-only policies, image-based enhancement baselines, and the proposed E-VLA models (Pick-Place results in Tab.~\ref{tab:1_main_results}). Under well-lit settings, event integration does not hurt performance: at 75\,lux all methods except E2VID reach 100\%, and at 40\,lux both overlay fusion and the event adapter achieve 100\% (vs.\ 80\% image-only).

\begin{figure}[t]
    \centering
    \includegraphics[width=\linewidth]{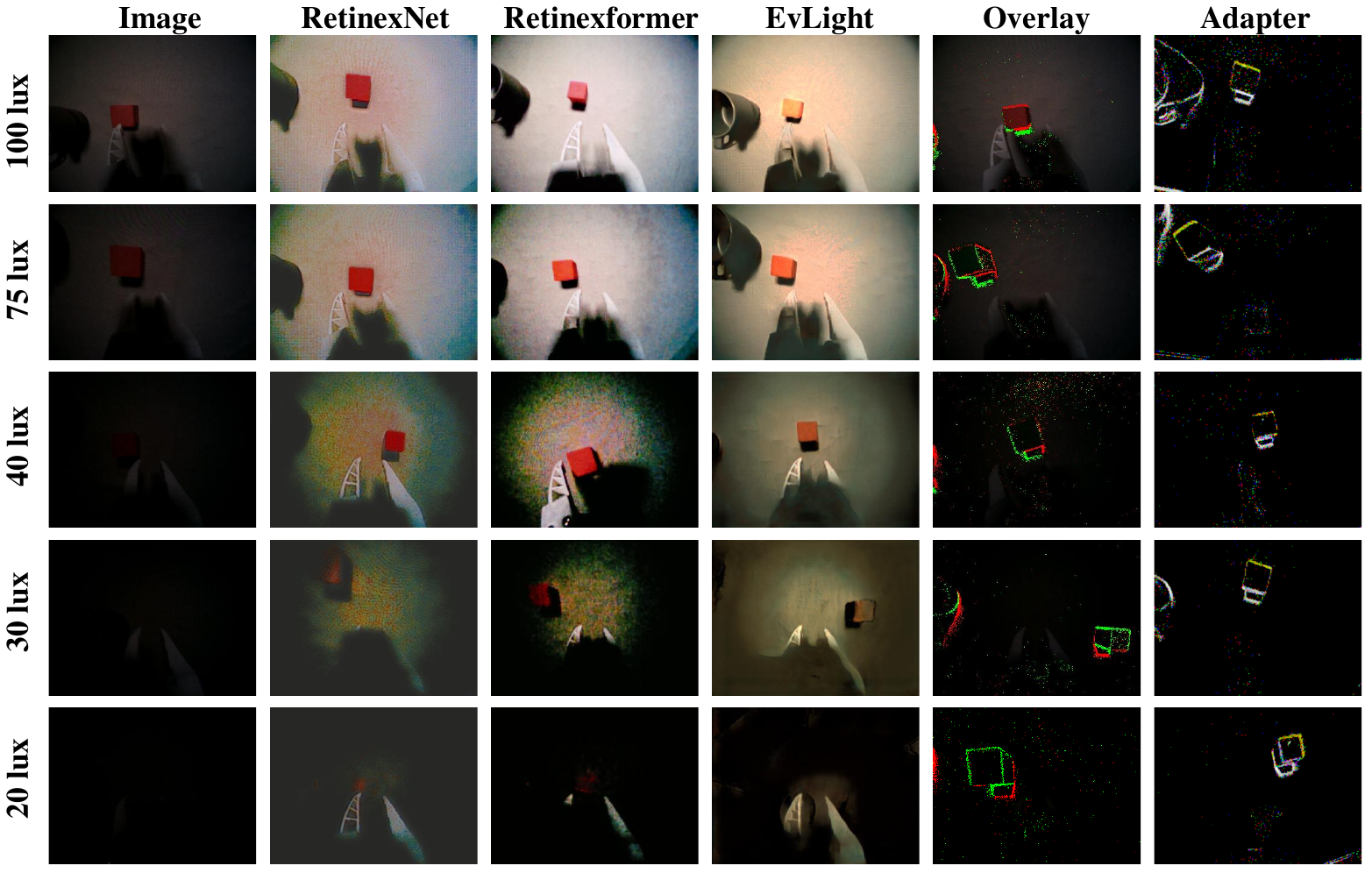}
    \vskip-2ex
    \caption{Qualitative comparison of visual inputs under different illumination.}
    \label{fig:input_vis}
    \vskip-4ex
\end{figure}

As illumination drops, the image-only baseline degrades from 70\% (35\,lux) to 35\% (30\,lux) and collapses to 0\% (25/20\,lux). Image enhancement helps only moderately (e.g., RetinexNet reaches 80\% at 30\,lux) but deteriorates at lower light (20--45\% at 25\,lux and 5--10\% at 20\,lux for several baselines), consistent with Fig.~\ref{fig:input_vis}. In contrast, E-VLA is markedly more robust: overlay achieves 75\% at 30\,lux and 65\%/60\% at 25/20\,lux, while the event adapter reaches 90\% at 30\,lux and maintains 90\% at 25/20\,lux. When frames are nearly black-clipped (25/20\,lux), image-only fails (0\%), whereas event-based variants retain substantial success (65--60\% overlay; 90\% adapter), showing that event streams provide actionable cues even when frames are uninformative. At even lower illumination, performance only drops near $\sim$2\,lux (Tab.~S1, supplementary material), where event noise begins to dominate.

\subsection{Generalization without Low-light Training}
% version with ID column
\begin{table}[t]
  \centering
  \footnotesize
  \setlength{\tabcolsep}{2pt}
  \renewcommand{\arraystretch}{1.25}
  \caption{Task success rates (\%) for Out-Of-Distribution (OOD) low-light conditions on the Pick-Place task. 
  In-Distribution (ID) refers to evaluating on low-light scenes that are included in training, for which the success rates are reported in Tab.~\ref{tab:1_main_results}. 
  OOD refers to evaluation on unseen low-light scenes.}
  \vskip-2ex
  \label{tab:2_ood}
  \resizebox{\textwidth}{!}{%
    \begin{tabular}{@{} l *{10}c S[table-format=2.1]}
      \toprule
      \rowcolor{tblhead}
      \multicolumn{1}{c}{} &
      \multicolumn{2}{c}{\textbf{100~lux}} &
      \multicolumn{2}{c}{\textbf{75~lux}} &
      \multicolumn{2}{c}{\textbf{40~lux}} &
      \multicolumn{2}{c}{\textbf{30~lux}} &
      \multicolumn{2}{c}{\textbf{20~lux}} &
      \multicolumn{1}{c}{\textbf{Average}} \\
      \rowcolor{tblhead}
      \multicolumn{1}{c}{\textbf{Method}} &
      \multicolumn{1}{c}{ID} & \multicolumn{1}{c}{OOD} &
      \multicolumn{1}{c}{ID} & \multicolumn{1}{c}{OOD} &
      \multicolumn{1}{c}{ID} & \multicolumn{1}{c}{OOD} &
      \multicolumn{1}{c}{ID} & \multicolumn{1}{c}{OOD} &
      \multicolumn{1}{c}{ID} & \multicolumn{1}{c}{OOD} &
      \multicolumn{1}{c}{OOD} \\
      \cmidrule(lr){2-3} \cmidrule(lr){4-5} \cmidrule(lr){6-7}
      \cmidrule(lr){8-9} \cmidrule(lr){10-11} \cmidrule(lr){12-12}
      \multicolumn{1}{l}{Image (10ms exp.)}
      &\cellcolor{heat1}100 & \cellcolor{heat1}100 & \cellcolor{heat1}100 & \cellcolor{heat6}65 & \cellcolor{heat2}80 & \cellcolor{heat6}30 & \cellcolor{heat6}35 & \cellcolor{heat6}0 & \cellcolor{heat6}0 & \cellcolor{heat6}0 & 39.0 \\
      \multicolumn{1}{l}{Image+Retinexformer~\cite{cai2023retinexformer}}
      & \cellcolor{heat1}100 & \cellcolor{heat1}100 & \cellcolor{heat1}100 & \cellcolor{heat3}70 & \cellcolor{heat2}80 & \cellcolor{heat5}50 & \cellcolor{heat3}75 & \cellcolor{heat6}10 & \cellcolor{heat6}10 & \cellcolor{heat6}0 & 46.0 \\
      \multicolumn{1}{l}{Image+EvLight~\cite{liang2024evlight}}
      & \cellcolor{heat1}100 & \cellcolor{heat2}80 & \cellcolor{heat1}100 & \cellcolor{heat5}50 & \cellcolor{heat2}95 & \cellcolor{heat5}50 & \cellcolor{heat3}75 & \cellcolor{heat6}20 & \cellcolor{heat6}10 & \cellcolor{heat6}5 & 40.5 \\
      \rowcolor{oursrow}
      \multicolumn{1}{l}{Ours overlay}
      & \cellcolor{heat1}100 & \cellcolor{heat1}100 & \cellcolor{heat1}100 & \cellcolor{heat1}95 & \cellcolor{heat1}100 & \cellcolor{heat5}55 & \cellcolor{heat3}75 & \cellcolor{heat6}10 & \cellcolor{heat4}60 & \cellcolor{heat6}10 & \second{54.0} \\
      \rowcolor{oursrow}
      \multicolumn{1}{l}{Ours adapter}
      & \cellcolor{heat1}100 & \cellcolor{heat1}100 & \cellcolor{heat1}100 & \cellcolor{heat2}85 & \cellcolor{heat1}100 & \cellcolor{heat5}75 & \cellcolor{heat1}90 & \cellcolor{heat3}70 & \cellcolor{heat1}90 & \cellcolor{heat5}45 & \best{75.0} \\
      \bottomrule
    \end{tabular}%
  }
\vspace{6pt}
    \begin{minipage}{\textwidth}
        \centering 
        \tiny \textit{Success rates (\%):} \hspace{0.25em}
        \colorbox{heat1}{\phantom{xx}} $\ge 90$ \textbf{(Best)}\hspace{0.25em}
        \colorbox{heat2}{\phantom{xx}} $80 \sim 90$ \hspace{0.25em}
        \colorbox{heat3}{\phantom{xx}} $70 \sim 80$ \hspace{0.25em}
        \colorbox{heat4}{\phantom{xx}} $60 \sim 70$ \hspace{0.25em}
        \colorbox{heat5}{\phantom{xx}} $50 \sim 60$ \hspace{0.25em}
        \colorbox{heat6}{\phantom{xx}} $\le 50$ (Bad)
    \end{minipage}
\end{table}
We further test OOD illumination generalization by training all models only on well-lit (200\,lux) demonstrations and evaluating at lower lux (Tab.~\ref{tab:2_ood}). The image-only policy drops from 65\% (75\,lux) to 30\% (40\,lux) and collapses to 0\% (30/20\,lux). In contrast, the event adapter remains robust: 85\% (75\,lux), 75\% (40\,lux), and 70\%/45\% (30/20\,lux). This shows event-based perception improves robustness to unseen illumination shifts and reduces the need for extensive low-light data collection.

\subsection{Motion Blur Performance}
We evaluate robustness under increasing motion blur by jointly adjusting illumination and exposure time (Tab.~\ref{tab:3_motion_blur}). Under mild blur, methods are comparable. As blur increases, image-only performance drops sharply: at 300\,ms, overlay improves Pick-Place from 40\% to 90\% and Sorting from 52.5\% to 72.5\%, while the event adapter reaches 85\%/85\%. Under severe blur (1000\,ms), image-only falls to 0\% (Pick-Place) and 5\% (Sorting), whereas overlay achieves 20\%/32.5\% and the event adapter further improves Pick-Place to 25\% (Sorting: 32.5\%). All models are trained without motion-blurred data, indicating gains stem from event sensing (high temporal resolution + recent-count windowing) rather than blur-specific adaptation, and mitigating the low-light/exposure trade-off of frame imaging.
\begin{table}[h]
  \centering
  \footnotesize
  \setlength{\tabcolsep}{2pt}
  \renewcommand{\arraystretch}{1.25}
  \caption{Task success rates (\%). Higher is better. Results are reported under different motion blur levels (represented by exposure time) on Pick-Place and Sorting task.}
  \vskip-2ex
  \label{tab:3_motion_blur}
  \resizebox{\textwidth}{!}{%
    \begin{tabular}{@{} l c S[table-format=2.1] c S[table-format=2.1] c S[table-format=2.1] c S[table-format=2.1] c S[table-format=2.1] c S[table-format=2.1] *{2}{S[table-format=2.1]} }
      \toprule
      \rowcolor{tblhead}
      \multicolumn{1}{c}{} &
      \multicolumn{2}{c}{\textbf{100~ms}} &
      \multicolumn{2}{c}{\textbf{200~ms}} &
      \multicolumn{2}{c}{\textbf{300~ms}} &
      \multicolumn{2}{c}{\textbf{400~ms}} &
      \multicolumn{2}{c}{\textbf{500~ms}} &
      \multicolumn{2}{c}{\textbf{1000~ms}} &
      \multicolumn{2}{c}{\textbf{Average}} \\
      \rowcolor{tblhead}
      \multicolumn{1}{c}{\textbf{Method}} &
      \multicolumn{1}{c}{P\&P} & \multicolumn{1}{c}{Sorting} &
      \multicolumn{1}{c}{P\&P} & \multicolumn{1}{c}{Sorting} &
      \multicolumn{1}{c}{P\&P} & \multicolumn{1}{c}{Sorting} &
      \multicolumn{1}{c}{P\&P} & \multicolumn{1}{c}{Sorting} &
      \multicolumn{1}{c}{P\&P} & \multicolumn{1}{c}{Sorting} &
      \multicolumn{1}{c}{P\&P} & \multicolumn{1}{c}{Sorting} &
      \multicolumn{1}{c}{P\&P} & \multicolumn{1}{c}{Sorting} \\
      \cmidrule(l){2-3} \cmidrule(l){4-5} \cmidrule(l){6-7}
      \cmidrule(l){8-9} \cmidrule(l){10-11} \cmidrule(l){12-13} 
      \cmidrule(l){14-15}
      \multicolumn{1}{l}{Image}
      & \cellcolor{heat1}100 & \cellcolor{heat3}77.5 & \cellcolor{heat2}85 & \cellcolor{heat3}75 & \cellcolor{heat6}40 & \cellcolor{heat5}52.5 & \cellcolor{heat6}30 & \cellcolor{heat6}32.5 & \cellcolor{heat6}15 & \cellcolor{heat6}20 & \cellcolor{heat6}0 & \cellcolor{heat6}5 & 45 & 43.75\\
      \rowcolor{oursrow}
      \multicolumn{1}{l}{Ours overlay}
      & \cellcolor{heat1}100 & \cellcolor{heat1}95 & \cellcolor{heat1}100 & \cellcolor{heat1}95 & \cellcolor{heat1}90 & \cellcolor{heat3}72.5 & \cellcolor{heat4}60 & \cellcolor{heat3}70 & \cellcolor{heat4}60 & \cellcolor{heat5}55 & \cellcolor{heat6}20 & \cellcolor{heat6}32.5 & \second{71.7} & \best{70.0}\\
      \rowcolor{oursrow}
      \multicolumn{1}{l}{Ours adapter}
      & \cellcolor{heat1}100 & \cellcolor{heat1}95 & \cellcolor{heat1}100 & \cellcolor{heat2}85 & \cellcolor{heat2}85 & \cellcolor{heat2}85 & \cellcolor{heat2}85 & \cellcolor{heat3}72.5 & \cellcolor{heat5}50 & \cellcolor{heat6}45 & \cellcolor{heat6}25 & \cellcolor{heat6}32.5 & \best{74.2} & \second{69.2} \\
      \bottomrule
    \end{tabular}%
  }
\vspace{6pt}
    \begin{minipage}{\textwidth}
        \centering 
        \tiny \textit{Success rates (\%):} \hspace{0.25em}
        \colorbox{heat1}{\phantom{xx}} $\ge 90$ (\textbf{Best}) \hspace{0.25em}
        \colorbox{heat2}{\phantom{xx}} $80 \sim 90$ \hspace{0.25em}
        \colorbox{heat3}{\phantom{xx}} $70 \sim 80$ \hspace{0.25em}
        \colorbox{heat4}{\phantom{xx}} $60 \sim 70$ \hspace{0.25em}
        \colorbox{heat5}{\phantom{xx}} $50 \sim 60$ \hspace{0.25em}
        \colorbox{heat6}{\phantom{xx}} $\le 50$ (Bad)
    \end{minipage}
\vskip-4ex
\end{table}

\subsection{Ablation Studies}

\begin{table}[t]
  \centering
  \footnotesize
  \setlength{\tabcolsep}{4pt}
  \renewcommand{\arraystretch}{1.12}
  \caption{Task success rates (\%) for different event windows. Higher is better. Results are reported under different illumination levels (lux) on the Pick-Place task.}
  \vskip-2ex
  \label{tab:4_event_window}
  \begin{tabular}{@{} l *{6}c S[table-format=2.1] @{}}
    \toprule
    \rowcolor{tblhead}
    \multicolumn{1}{l}{\textbf{Event Window}} &
    \multicolumn{1}{c}{\textbf{75~lux}} &
    \multicolumn{1}{c}{\textbf{40~lux}} &
    \multicolumn{1}{c}{\textbf{35~lux}} &
    \multicolumn{1}{c}{\textbf{30~lux}} &
    \multicolumn{1}{c}{\textbf{25~lux}} &
    \multicolumn{1}{c}{\textbf{20~lux}} &
    \multicolumn{1}{c}{\textbf{Average}} \\
    \midrule
    \midrule
    % ---- time windows ----
    \multicolumn{1}{l}{5\,ms window} & \cellcolor{heat2}80 & \cellcolor{heat3}70 & \cellcolor{heat4}65 & \cellcolor{heat5}55 & \cellcolor{heat5}50 & \cellcolor{heat6}45 & 60.8\\
    \multicolumn{1}{l}{20\,ms window} & \cellcolor{heat1}100 & \cellcolor{heat2}85 & \cellcolor{heat2}85 & \cellcolor{heat2}80 & \cellcolor{heat3}75 & \cellcolor{heat3}70 & 82.5\\
    \multicolumn{1}{l}{40\,ms window} & \cellcolor{heat1}100 & \cellcolor{heat1}90 & \cellcolor{heat1}90 & \cellcolor{heat3}75 & \cellcolor{heat5}50 & \cellcolor{heat6}45 & 75\\
    \addlinespace[2pt]
    \midrule
    \addlinespace[2pt]
    % ---- event count windows ----
    \multicolumn{1}{l}{500 events} & \cellcolor{heat1}100 & \cellcolor{heat1}100 & \cellcolor{heat4}60 & \cellcolor{heat5}50 & \cellcolor{heat5}50 & \cellcolor{heat6}45 & 67.5 \\
    \multicolumn{1}{l}{2000 events} & \cellcolor{heat1}100 & \cellcolor{heat1}100 & \cellcolor{heat1}95 & \cellcolor{heat1}90 & \cellcolor{heat1}90 & \cellcolor{heat1}90 & \best{94.2} \\
    \multicolumn{1}{l}{4000 events} & \cellcolor{heat1}100 & \cellcolor{heat1}100 & \cellcolor{heat2}85 & \cellcolor{heat2}85 & \cellcolor{heat3}75 & \cellcolor{heat3}70 & \second{85.8} \\
    \bottomrule
  \end{tabular}%
  \vspace{6pt}
    \begin{minipage}{\textwidth}
        \centering 
        \tiny \textit{Success rates (\%):} \hspace{0.25em}
        \colorbox{heat1}{\phantom{xx}} $\ge 90$ (\textbf{Best}) \hspace{0.25em}
        \colorbox{heat2}{\phantom{xx}} $80 \sim 90$ \hspace{0.25em}
        \colorbox{heat3}{\phantom{xx}} $70 \sim 80$ \hspace{0.25em}
        \colorbox{heat4}{\phantom{xx}} $60 \sim 70$ \hspace{0.25em}
        \colorbox{heat5}{\phantom{xx}} $50 \sim 60$ \hspace{0.25em}
        \colorbox{heat6}{\phantom{xx}} $\le 50$ (Bad)
    \end{minipage}
  \vskip-2ex
\end{table}

We analyze key design choices for event-based VLA, including event windowing and training recipes.

\noindent \textbf{Event windowing strategy and size.} Tab.~\ref{tab:4_event_window} shows time-based windows (5/20/40\,ms) are unreliable under synchronous manipulation due to pauses and speed changes, while event-count windows are more stable. A moderate window performs best: 2000 events yields 95\% at 35\,lux and 90\% at 30/25/20\,lux, outperforming 500 events (60\% at 35\,lux; 45\% at 20\,lux) and 4000 events (85\% at 35\,lux; 70\% at 20\,lux).

\noindent \textbf{Training recipe and weight-sharing.} Tab.~\ref{tab:5_traing} confirms progressive training is critical: Action$\rightarrow$Joint achieves 75\%, while Event$\rightarrow$Joint drops to 50\%. Using both Action and Event stages improves to 80\% (Action$\rightarrow$Event$\rightarrow$Joint, no sharing), and the full recipe with patch-embedding weight-sharing reaches 90\%, supporting stable perception--action alignment and cross-modal integration.

\begin{table}[h]
  \centering
  \footnotesize
  \setlength{\tabcolsep}{4pt}
  \renewcommand{\arraystretch}{1.15}
  \caption{Task success rates (\%) on different training recipes at 30~lux. Higher is better. `sharing' represents weight-sharing for the patch embedding layer.}
  \vskip-2ex
  \label{tab:5_traing}
  \begin{tabular}{@{} l cccc S[table-format=2.1]}
    \toprule
    \rowcolor{tblhead}
    \multicolumn{1}{c}{\textbf{Strategy}} &
    \multicolumn{1}{c}{\textbf{Action}} &
    \multicolumn{1}{c}{\textbf{Event}} &
    \multicolumn{1}{c}{\textbf{Joint}} &
    \multicolumn{1}{c}{\textbf{Sharing}} &
    \multicolumn{1}{c}{\textbf{Success}} \\
    \midrule
    \midrule
    \multicolumn{1}{l}{Action$\rightarrow$Joint}     & \ding{51} & \ding{55} & \ding{51} & \ding{51} & 75.0 \\
    \multicolumn{1}{l}{Event$\rightarrow$Joint}         & \ding{55} & \ding{51} & \ding{51} & \ding{51} & 50.0 \\
    \multicolumn{1}{l}{Action $\rightarrow$ Event$\rightarrow$Joint}         & \ding{51} & \ding{51} & \ding{51}  & \ding{55} & \second{80.0} \\
    \rowcolor{oursrow}
    \multicolumn{1}{l}{Ours}         & \ding{51} & \ding{51} & \ding{51}  & \ding{51} & \best{90.0} \\
    \bottomrule
  \end{tabular}
  \vskip-4ex
\end{table}

\subsection{Failure Cases: Color Semantics and Occlusion}

We further investigate Sorting and Stacking, which introduce task-specific perceptual challenges beyond those in Pick-Place.

\noindent \textbf{1) Color-based sorting.} 
Unlike the Pick-Place task, the Sorting task additionally requires correct color-based placement. Although grasping remains reliable, failures mainly occur during placement due to the lower color classification accuracy of event-based methods when RGB frames are degraded. This limitation is inherent to event sensing: events encode intensity changes rather than absolute appearance, static object color vanishes after grasp stabilization, and sparse RGB events do not satisfy dense demosaicing assumptions. It suggests that, while event cues are effective for maintaining object-level perception under low illumination, tasks that require reliable color discrimination remain supported by intensity-based visual signals.
\noindent \textbf{2) Stacking with occlusion.} 
In the Stacking task, failures primarily arise during the stack phase due to the viewpoint occlusion caused by the grasped object blocking the wrist-mounted camera~\cite{ma2024survey}. 
Therefore, the availability of event information does not substantially alter the perceptual bottleneck.
Overall, these results indicate that E-VLA is effective for low-light manipulation tasks. Limitations observed in color-driven behaviors point to representation-level challenges that may be addressed with more advanced event-based color reconstruction, whereas failures under strong viewpoint occlusion reflect perceptual constraints that are not directly mitigated by event sensing. More detailed success rate analyses can be found in the supplementary material.
\section{Conclusions}
\label{sec:conc}

We present E-VLA, an event-augmented vision-language-action (VLA) framework that improves VLA robustness under visual degradations, including extreme low light, black clipping, and motion blur. Instead of reconstructing images from events, E-VLA integrates asynchronous event streams into a pretrained VLA architecture in a lightweight, token-compatible manner, enabling robust manipulation without large-scale event pretraining.
To facilitate systematic study, we build an open-source teleoperation platform with a DAVIS346 event camera and collect a synchronized RGB--event--action manipulation dataset spanning tasks, illumination levels, and motion regimes. Experiments show that even a parameter-free fusion (\ie, overlaying accumulated event maps onto RGB images) improves success when frame inputs become unreliable, while preserving performance in bright scenes. On Pick-Place at 20\,lux, success increases from 0\% (image-only) to 60\% with overlay fusion and to 90\% with our event adapter; under severe motion blur (1000\,ms exposure), Pick-Place improves from 0\% to 20--25\%, and Sorting from 5\% to 32.5\%. Moreover, E-VLA improves out-of-distribution generalization across illumination shifts and blur-heavy dynamics.
Our ablations provide design insights for coupling event perception with vision-pretrained policy models, highlighting event windowing/accumulation, representation choice, and training balance for stable perception--action alignment. We hope this work encourages adoption of event-driven sensing in VLA learning and inspires robust embodied intelligence beyond frame-based imaging.

\section*{Acknowledgments}
This research was funded by the Natural Science Foundation of Zhejiang Province (Grant No. LZ24F050003), the National Natural Science Foundation of China (Grant No. 62473139), the Hunan Provincial Research and Development Project (Grant No. 2025QK3019), and the opening project of the State Key Laboratory of Autonomous Intelligent Unmanned Systems (Grant No. ZZKF2025-2-10).

\bibliographystyle{splncs04}
\bibliography{main}

\begin{thebibliography}{10}
\providecommand{\url}[1]{\texttt{#1}}
\providecommand{\urlprefix}{URL }
\providecommand{\doi}[1]{https://doi.org/#1}

\bibitem{alonso2019ev_segnet}
Alonso, I., Murillo, A.C.: {EV-SegNet:} {Semantic} segmentation for event-based cameras. In: CVPRW (2019)

\bibitem{bao2024temporal}
Bao, Y., Sun, L., Ma, Y., Wang, K.: Temporal-mapping photography for event cameras. In: ECCV (2024)

\bibitem{bi2025vlatouch}
Bi, J., Ma, K.Y., Hao, C., Shou, M.Z., Soh, H.: {VLA-Touch:} {Enhancing} vision-language-action models with dual-level tactile feedback. arXiv preprint arXiv:2507.17294  (2025)

\bibitem{black2024pi_0}
Black, K., Brown, N., Driess, D., Esmail, A., Equi, M., Finn, C., Fusai, N., Groom, L., Hausman, K., Ichter, B., Jakubczak, S., Jones, T., Ke, L., Levine, S., Li{-}Bell, A., Mothukuri, M., Nair, S., Pertsch, K., Shi, L.X., Tanner, J., Vuong, Q., Walling, A., Wang, H., Zhilinsky, U.: {\(\pi\)}\({}_{\mbox{0}}\): {A} vision-language-action flow model for general robot control. arXiv preprint arXiv:2410.24164  (2024)

\bibitem{brohan2022rt1}
Brohan, A., Brown, N., Carbajal, J., Chebotar, Y., Dabis, J., Finn, C., Gopalakrishnan, K., Hausman, K., Herzog, A., Hsu, J., Ibarz, J., Ichter, B., Irpan, A., Jackson, T., Jesmonth, S., Joshi, N.J., Julian, R., Kalashnikov, D., Kuang, Y., Leal, I., Lee, K., Levine, S., Lu, Y., Malla, U., Manjunath, D., Mordatch, I., Nachum, O., Parada, C., Peralta, J., Perez, E., Pertsch, K., Quiambao, J., Rao, K., Ryoo, M.S., Salazar, G., Sanketi, P.R., Sayed, K., Singh, J., Sontakke, S., Stone, A., Tan, C., Tran, H.T., Vanhoucke, V., Vega, S., Vuong, Q., Xia, F., Xiao, T., Xu, P., Xu, S., Yu, T., Zitkovich, B.: {RT-1:} {Robotics} transformer for real-world control at scale. In: RSS (2022)

\bibitem{bugueno2025humanrobotnav}
Bugue{\~{n}}o{-}C{\'{o}}rdova, I.G., Ruiz{-}del{-}Solar, J., Verschae, R.: Human-robot navigation using event-based cameras and reinforcement learning. In: CVPRW (2025)

\bibitem{cadena2021spade-e2vid}
Cadena, P.R.G., Qian, Y., Wang, C., Yang, M.: {SPADE-E2VID:} {Spatially-adaptive} denormalization for event-based video reconstruction. IEEE Transactions on Image Processing  (2021)

\bibitem{cadene2024lerobot}
Cadene, R., Alibert, S., Soare, A., Gallouedec, Q., Zouitine, A., Palma, S., Kooijmans, P., Aractingi, M., Shukor, M., Aubakirova, D., Russi, M., Capuano, F., Pascal, C., Choghari, J., Moss, J., Wolf, T.: {LeRobot:} {State-of-the-art} machine learning for real-world robotics in pytorch. \url{https://github.com/huggingface/lerobot} (2024)

\bibitem{cai2023retinexformer}
Cai, Y., Bian, H., Lin, J., Wang, H., Timofte, R., Zhang, Y.: {Retinexformer:} {One-stage} retinex-based transformer for low-light image enhancement. In: ICCV (2023)

\bibitem{cao2024chasing}
Cao, J., Zheng, X., Lyu, Y., Wang, J., Xu, R., Wang, L.: Chasing day and night: Towards robust and efficient all-day object detection guided by an event camera. In: ICRA (2024)

\bibitem{chen2025evlightpp}
Chen, K., Liang, G., Lu, Y., Li, H., Wang, L.: {EvLight++:} {Low-light} video enhancement with an event camera: A large-scale real-world dataset, novel method, and more. IEEE Transactions on Pattern Analysis and Machine Intelligence  (2026)

\bibitem{chen2025tgrpo}
Chen, Z., Niu, R., Kong, H., Wang, Q., Xing, Q., Fan, Z.: {TGRPO:} {Fine-tuning} vision-language-action model via trajectory-wise group relative policy optimization. arXiv preprint arXiv:2506.08440  (2025)

\bibitem{cheng2025omnivtla}
Cheng, Z., Zhang, Y., Zhang, W., Li, H., Wang, K., Song, L., Zhang, H.: {OmniVTLA:} {Vision}-tactile-language-action model with semantic-aligned tactile sensing. arXiv preprint arXiv:2508.08706  (2025)

\bibitem{delbruck2008framefree}
Delbr{\"u}ck, T.: Frame-free dynamic digital vision. In: International Symposium on Secure-Life Electronics, Advanced Electronics for Quality Life and Society (2008)

\bibitem{deng2025SurveyRL}
Deng, H., Wu, Z., Liu, H., Guo, W., Xue, Y., Shan, Z., Zhang, C., Jia, B., Ling, Y., Lu, G.: A survey on reinforcement learning of vision-language-action models for robotic manipulation. Authorea Preprints  (2025)

\bibitem{deng2025stereovla}
Deng, S., Yan, M., Zheng, Y., Su, J., Zhang, W., Zhao, X., Cui, H., Zhang, Z., Wang, H.: {StereoVLA:} {Enhancing} vision-language-action models with stereo vision. arXiv preprint arXiv:2512.21970  (2025)

\bibitem{libero-plus2025}
Fei, S., Wang, S., Shi, J., Dai, Z., Cai, J., Qian, P., Ji, L., He, X., Zhang, S., Fei, Z., Fu, J., Gong, J., Qiu, X.: {LIBERO-Plus:} {In-depth} robustness analysis of vision-language-action models. arXiv preprint arXiv:2510.13626  (2025)

\bibitem{funk2024evetac}
Funk, N., Helmut, E., Chalvatzaki, G., Calandra, R., Peters, J.: Evetac: An event-based optical tactile sensor for robotic manipulation. IEEE Transactions on Robotics  (2024)

\bibitem{gallego2020eventsurvey}
Gallego, G., Delbr{\"{u}}ck, T., Orchard, G., Bartolozzi, C., Taba, B., Censi, A., Leutenegger, S., Davison, A.J., Conradt, J., Daniilidis, K., Scaramuzza, D.: Event-based vision: A survey. IEEE Transactions on Pattern Analysis and Machine Intelligence  (2022)

\bibitem{gehrig2021dsec}
Gehrig, M., Aarents, W., Gehrig, D., Scaramuzza, D.: {DSEC:} {A} stereo event camera dataset for driving scenarios. IEEE Robotics and Automation Letters  (2021)

\bibitem{gehrig2021eraft}
Gehrig, M., Millh{\"a}usler, M., Gehrig, D., Scaramuzza, D.: {E-RAFT:} {Dense} optical flow from event cameras. In: 3DV (2021)

\bibitem{guo2024force}
Guo, Q., Yu, Z., Fu, J., Lu, Y., Zweiri, Y., Gan, D.: {Force-EvT:} {A} closer look at robotic gripper force measurement with event-based vision transformer. In: ReMAR (2024)

\bibitem{guo2025irevla}
Guo, Y., Zhang, J., Chen, X., Ji, X., Wang, Y.J., Hu, Y., Chen, J.: Improving vision-language-action model with online reinforcement learning. arXiv preprint arXiv:2501.16664  (2025)

\bibitem{huang2025tactile}
Huang, J., Wang, S., Lin, F., Hu, Y., Wen, C., Gao, Y.: {Tactile-VLA:} {Unlocking} vision-language-action model's physical knowledge for tactile generalization. arXiv preprint arXiv:2507.09160  (2025)

\bibitem{huang2022visiongrasp}
Huang, X., Halwani, M., Muthusamy, R., Ayyad, A., Swart, D., Seneviratne, L., Gan, D., Zweiri, Y.: Real-time grasping strategies using event camera. Journal of Intelligent Manufacturing  (2022)

\bibitem{intelligence2025pi_0.6}
Intelligence, P., Amin, A., Aniceto, R., Balakrishna, A., Black, K., Conley, K., Connors, G., Darpinian, J., Dhabalia, K., DiCarlo, J., Driess, D., Equi, M., Esmail, A., Fang, Y., Finn, C., Glossop, C., Godden, T., Goryachev, I., Groom, L., Hancock, H., Hausman, K., Hussein, G., Ichter, B., Jakubczak, S., Jen, R., Jones, T., Katz, B., Ke, L., Kuchi, C., Lamb, M., LeBlanc, D., Levine, S., Li{-}Bell, A., Lu, Y., Mano, V., Mothukuri, M., Nair, S., Pertsch, K., Ren, A.Z., Sharma, C., Shi, L.X., Smith, L., Springenberg, J.T., Stachowicz, K., Stoeckle, W., Swerdlow, A., Tanner, J., Torne, M., Vuong, Q., Walling, A., Wang, H., Williams, B., Yoo, S., Yu, L., Zhilinsky, U., Zhou, Z.: $\pi^{*}_{0.6}$: a {VLA} that learns from experience. arXiv preprint arXiv:2511.14759  (2025)

\bibitem{intelligence2025pi_0.5}
Intelligence, P., Black, K., Brown, N., Darpinian, J., Dhabalia, K., Driess, D., Esmail, A., Equi, M., Finn, C., Fusai, N., Galliker, M.Y., Ghosh, D., Groom, L., Hausman, K., Ichter, B., Jakubczak, S., Jones, T., Ke, L., LeBlanc, D., Levine, S., Li{-}Bell, A., Mothukuri, M., Nair, S., Pertsch, K., Ren, A.Z., Shi, L.X., Smith, L., Springenberg, J.T., Stachowicz, K., Tanner, J., Vuong, Q., Walke, H., Walling, A., Wang, H., Yu, L., Zhilinsky, U.: {\(\pi\)}\({}_{\mbox{0.5}}\): a vision-language-action model with open-world generalization. arXiv preprint arXiv:2504.16054  (2025)

\bibitem{kim2024openvla}
Kim, M.J., Pertsch, K., Karamcheti, S., Xiao, T., Balakrishna, A., Nair, S., Rafailov, R., Foster, E.P., Sanketi, P.R., Vuong, Q., Kollar, T., Burchfiel, B., Tedrake, R., Sadigh, D., Levine, S., Liang, P., Finn, C.: {OpenVLA:} {An} open-source vision-language-action model. In: CoRL (2024)

\bibitem{lagorce2016timesurface}
Lagorce, X., Orchard, G., Galluppi, F., Shi, B.E., Benosman, R.: {HOTS:} {A} hierarchy of event-based time-surfaces for pattern recognition. IEEE Transactions on Pattern Analysis and Machine Intelligence  (2017)

\bibitem{li2020eventgrasp}
Li, B., Cao, H., Qu, Z., Hu, Y., Wang, Z., Liang, Z.: Event-based robotic grasping detection with neuromorphic vision sensor and event-grasping dataset. Frontiers in Neurorobotics  (2020)

\bibitem{li2025pointvla}
Li, C., Wen, J., Peng, Y., Peng, Y., Feng, F., Zhu, Y.: {PointVLA:} {Injecting} the {3D} world into vision-language-action models. arXiv preprint arXiv:2503.07511  (2025)

\bibitem{li2025survey}
Li, H., Chen, Y., Cui, W., Liu, W., Liu, K., Zhou, M., Zhang, Z., Zhao, D.: Survey of vision-language-action models for embodied manipulation. arXiv preprint arXiv:2508.15201  (2025)

\bibitem{li2024eventseg}
Li, H., Wang, J., Yuan, J., Li, Y., Weng, W., Peng, Y., Zhang, Y., Xiong, Z., Sun, X.: Event-assisted low-light video object segmentation. In: CVPR (2024)

\bibitem{li2024flowtrackingdataset}
Li, Y., Shen, Y., Huang, Z., Chen, S., Bian, W., Shi, X., Wang, F.Y., Sun, K., Bao, H., Cui, Z., Zhang, G., Li, H.: {BlinkVision:} {A} benchmark for optical flow, scene flow and point tracking estimation using rgb frames and events. In: ECCV (2024)

\bibitem{liang2024evlight}
Liang, G., Chen, K., Li, H., Lu, Y., Wang, L.: Towards robust event-guided low-light image enhancement: {A} large-scale real-world event-image dataset and novel approach. In: CVPR (2024)

\bibitem{liu2025tracking}
Liu, J., Wang, B., Tan, Z., Zhang, J., Shen, H., Hu, D.: Tracking any point with frame-event fusion network at high frame rate. In: IROS (2025)

\bibitem{liu2024rdt}
Liu, S., Wu, L., Li, B., Tan, H., Chen, H., Wang, Z., Xu, K., Su, H., Zhu, J.: {RDT-1B:} {a} diffusion foundation model for bimanual manipulation. arXiv preprint arXiv:2410.07864  (2024)

\bibitem{lu2025vlarl}
Lu, G., Guo, W., Zhang, C., Zhou, Y., Jiang, H., Gao, Z., Tang, Y., Wang, Z.: {VLA-RL:} {Towards} masterful and general robotic manipulation with scalable reinforcement learning. arXiv preprint arXiv:2505.18719  (2025)

\bibitem{lu2025uniinr}
Lu, Y., Liang, G., Wang, Y., Wang, L., Xiong, H.: {UniINR:} {Event-guided} unified rolling shutter correction, deblurring, and interpolation. In: ECCV (2025)

\bibitem{color4e}
Ma, Y., Duan, P., Hong, Y., Zhou, C., Zhang, Y., Ren, J., Shi, B.: Color4e: Event demosaicing for full-color event guided image deblurring. In: ACMMM (2024)

\bibitem{ma2024survey}
Ma, Y., Song, Z., Zhuang, Y., Hao, J., King, I.: A survey on vision-language-action models for embodied {AI}. arXiv preprint arXiv:2405.14093  (2024)

\bibitem{marafioti2025smolvlm}
Marafioti, A., Zohar, O., Farr{\'e}, M., Noyan, M., Bakouch, E., Cuenca, P., Zakka, C., Allal, L.B., Lozhkov, A., Tazi, N., Srivastav, V., Lochner, J., Larcher, H., Morlon, M., Tunstall, L., von Werra, L., Wolf, T.: {SmolVLM:} {Redefining} small and efficient multimodal models. arXiv preprint arXiv:2504.05299  (2025)

\bibitem{muthusamy2021neurograsp}
Muthusamy, R., Ayyad, A., Halwani, M., Swart, D., Gan, D., Seneviratne, L., Zweiri, Y.: Neuromorphic eye-in-hand visual servoing. IEEE Access  (2021)

\bibitem{nah2017gopro}
Nah, S., Hyun~Kim, T., Mu~Lee, K.: Deep multi-scale convolutional neural network for dynamic scene deblurring. In: CVPR (2017)

\bibitem{pan2019edi}
Pan, L., Scheerlinck, C., Yu, X., Hartley, R., Liu, M., Dai, Y.: Bringing a blurry frame alive at high frame-rate with an event camera. In: CVPR (2019)

\bibitem{rebecq2019e2vid}
Rebecq, H., Ranftl, R., Koltun, V., Scaramuzza, D.: Events-to-video: Bringing modern computer vision to event cameras. In: CVPR (2019)

\bibitem{reinold2025simuslip}
Reinold, T., Ghosh, S., Gallego, G.: Combined physics and event camera simulator for slip detection. In: WACVW (2025)

\bibitem{neuromorphicnav2025}
Sanyal, S., Joshi, A., Kosta, A., Roy, K.: Real-time neuromorphic navigation: Guiding physical robots with event-based sensing and task-specific reconfigurable autonomy stack (2025)

\bibitem{shang2021d2net}
Shang, W., Ren, D., Zou, D., Ren, J.S., Luo, P., Zuo, W.: Bringing events into video deblurring with non-consecutively blurry frames. In: ICCV (2021)

\bibitem{shukor2025smolvla}
Shukor, M., Aubakirova, D., Capuano, F., Kooijmans, P., Palma, S., Zouitine, A., Aractingi, M., Pascal, C., Russi, M., Marafioti, A., Alibert, S., Cord, M., Wolf, T., Cad{\`{e}}ne, R.: {SmolVLA:} {A} vision-language-action model for affordable and efficient robotics. arXiv preprint arXiv:2506.01844  (2025)

\bibitem{singh2025ogvla}
Singh, I., Goyal, A., Birchfield, S., Fox, D., Garg, A., Blukis, V.: {OG-VLA:} {3D}-aware vision language action model via orthographic image generation. arXiv preprint arXiv:2506.01196  (2025)

\bibitem{stoffregen2020hqf}
Stoffregen, T., Scheerlinck, C., Scaramuzza, D., Drummond, T., Barnes, N., Kleeman, L., Mahony, R.E.: Reducing the sim-to-real gap for event cameras. In: ECCV (2020)

\bibitem{sun2025lowlight}
Sun, L., Bao, Y., Zhai, J., Liang, J., Zhang, Y., Wang, K., Paudel, D.P., Van~Gool, L.: Low-light image enhancement using event-based illumination estimation. In: ICCV (2025)

\bibitem{sun2022efnet}
Sun, L., Sakaridis, C., Liang, J., Jiang, Q., Yang, K., Sun, P., Ye, Y., Wang, K., Gool, L.V.: Event-based fusion for motion deblurring with cross-modal attention. In: ECCV (2022)

\bibitem{sun2023highrev}
Sun, L., Sakaridis, C., Liang, J., Sun, P., Zhang, K., Cao, J., Jiang, Q., Wang, K., Van~Gool, L.: Event-based frame interpolation with ad-hoc deblurring. In: CVPR (2023)

\bibitem{sun2025geovla}
Sun, L., Xie, B., Liu, Y., Shi, H., Wang, T., Cao, J.: {GeoVLA:} {Empowering} {3D} representations in vision-language-action models. arXiv preprint arXiv:2508.09071  (2025)

\bibitem{sun2024motion_representation}
Sun, Z., Fu, X., Huang, L., Liu, A., Zha, Z.J.: Motion aware event representation-driven image deblurring. In: ECCV (2024)

\bibitem{taunyazov2020eventtact}
Taunyazov, T., Sng, W., Lim, B., See, H., Kuan, J., Ansari, A.F., Tee, B.C.K., Soh, H.: Event-driven visual-tactile sensing and learning for robots. In: RSS (2020)

\bibitem{team2025gemini}
Team, G.R., Abeyruwan, S., Ainslie, J., Alayrac, J.B., Arenas, M.G., Armstrong, T., Balakrishna, A., Baruch, R., Bauza, M., Blokzijl, M., et~al.: Gemini robotics: Bringing {AI} into the physical world. arXiv preprint arXiv:2503.20020  (2025)

\bibitem{tomy2022fusing_adverse}
Tomy, A., Paigwar, A., Mann, K.S., Renzaglia, A., Laugier, C.: Fusing event-based and {RGB} camera for robust object detection in adverse conditions. In: ICRA (2022)

\bibitem{wang2024towards_robust_tracking}
Wang, X., Yu, H., Yu, L., Yang, W., Xia, G.S.: Towards robust keypoint detection and tracking: A fusion approach with event-aligned image features. IEEE Robotics and Automation Letters  (2024)

\bibitem{wang2025vlaadapter}
Wang, Y., Ding, P., Li, L., Cui, C., Ge, Z., Tong, X., Song, W., Zhao, H., Zhao, W., Hou, P., Huang, S., Tang, Y., Wang, W., Zhang, R., Liu, J., Wang, D.: {VLA-Adapter:} {An} effective paradigm for tiny-scale vision-language-action model. arXiv preprint arXiv:2509.09372  (2025)

\bibitem{wei2018retinexnet}
Wei, C., Wang, W., Yang, W., Liu, J.: Deep retinex decomposition for low-light enhancement. arXiv preprint arXiv:1808.04560  (2018)

\bibitem{wen2025tinyvla}
Wen, J., Zhu, Y., Li, J., Zhu, M., Wu, K., Xu, Z., Liu, N., Cheng, R., Shen, C., Peng, Y., Feng, F., Tang, J.: {TinyVLA:} {Towards} fast, data-efficient vision-language-action models for robotic manipulation. IEEE Robotics and Automation Letters  (2025)

\bibitem{ye2023towards_event}
Ye, Y., Shi, H., Yang, K., Wang, Z., Yin, X., Sun, L., Wang, Y., Wang, K.: Towards anytime optical flow estimation with event cameras. Sensors  (2025)

\bibitem{zhong2025survey}
Zhong, Y., Bai, F., Cai, S., Huang, X., Chen, Z., Zhang, X., Wang, Y., Guo, S., Guan, T., Lui, K.N., Qi, Z., Liang, Y., Chen, Y., Yang, Y.: A survey on vision-language-action models: An action tokenization perspective. arXiv preprint arXiv:2507.01925  (2025)

\bibitem{zihao2018unsupervised}
Zihao~Zhu, A., Yuan, L., Chaney, K., Daniilidis, K.: Unsupervised event-based optical flow using motion compensation. In: ECCVW (2018)

\bibitem{zitkovich2023rt2}
Zitkovich, B., Yu, T., Xu, S., Xu, P., Xiao, T., Xia, F., Wu, J., Wohlhart, P., Welker, S., Wahid, A., Vuong, Q., Vanhoucke, V., Tran, H.T., Soricut, R., Singh, A., Singh, J., Sermanet, P., Sanketi, P.R., Salazar, G., Ryoo, M.S., Reymann, K., Rao, K., Pertsch, K., Mordatch, I., Michalewski, H., Lu, Y., Levine, S., Lee, L., Lee, T.E., Leal, I., Kuang, Y., Kalashnikov, D., Julian, R., Joshi, N.J., Irpan, A., Ichter, B., Hsu, J., Herzog, A., Hausman, K., Gopalakrishnan, K., Fu, C., Florence, P., Finn, C., Dubey, K.A., Driess, D., Ding, T., Choromanski, K.M., Chen, X., Chebotar, Y., Carbajal, J., Brown, N., Brohan, A., Arenas, M.G., Han, K.: {RT-2:} {Vision}-language-action models transfer web knowledge to robotic control. In: CoRL (2023)

\end{thebibliography}

\renewcommand{\thetable}{S\arabic{table}}
\renewcommand{\thefigure}{S\arabic{figure}}

This document provides additional materials that complement the main paper. First, Sec.~\ref{Sec:imple_details} describes implementation details, including illumination control, dataset collection, and training procedures. Sec.~\ref{Sec:addi_exp} reports additional experiments under even lower illumination, together with a detailed analysis of the E2VID approach. Sec.~\ref{Sec:effi} compares the computational efficiency of all methods. Sec.~\ref{Sec:ablation} provides more ablation studies, and Sec.~\ref{Sec:mani_seq} presents example manipulation processes. Finally, we discuss the limitations of the proposed method and outline potential directions for improvement.

\section{Implementation Details}
\label{Sec:imple_details}

\subsection{Illumination Setting}

The illumination level is controlled using a tunable light source and measured at the camera location with a lux meter. Additional extremely low-light conditions below $20$~lux are produced by placing neutral density filters in front of the light source. Our main experiments are conducted under $20 \sim 75$~lux. Since the brightness of captured images depends not only on ambient illumination but also on sensor characteristics and the imaging pipeline, we characterize the effective lighting condition using the pixel intensity of captured RGB frames.

\begin{figure}[h]
    \centering
\includegraphics[width=\linewidth]{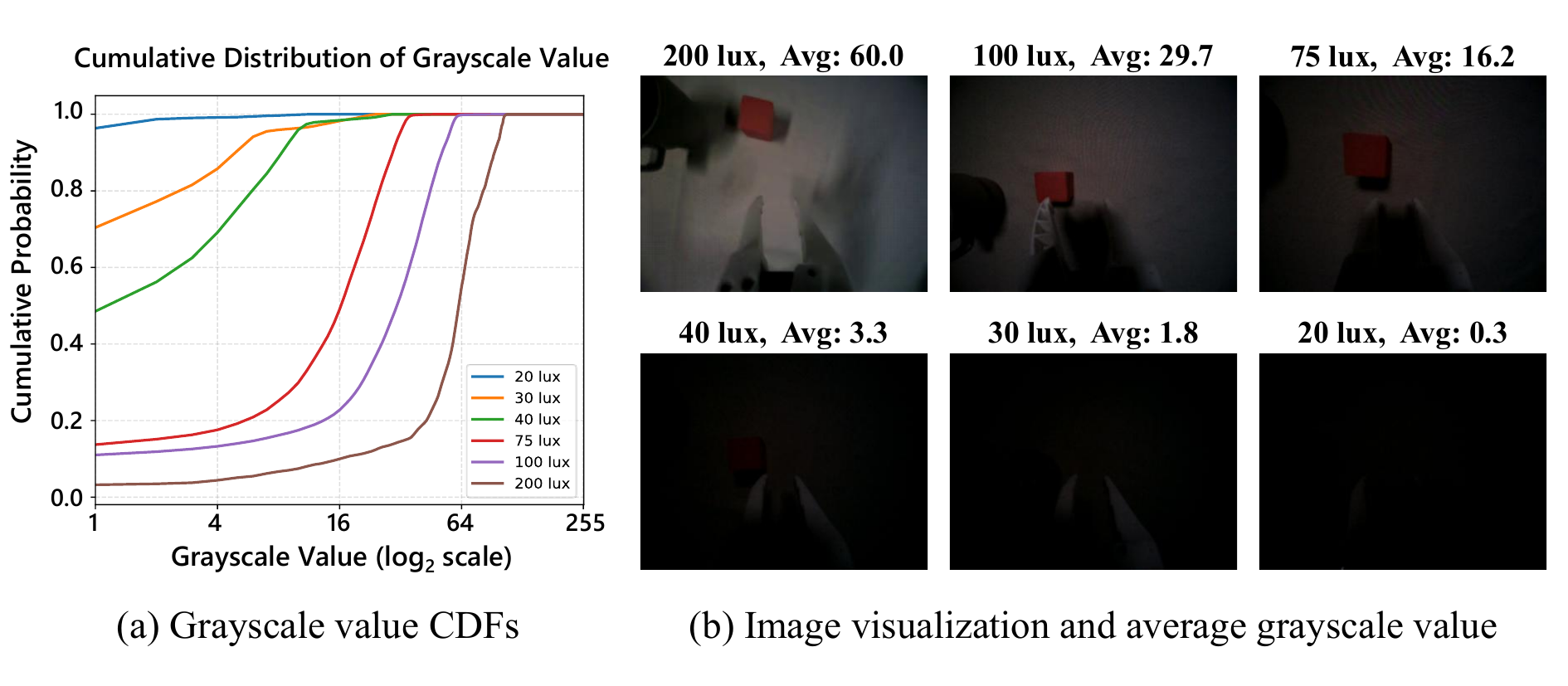}
    \vskip-2ex
    \caption{Grayscale distribution of images captured under different illumination levels. (a) Cumulative distribution functions (CDFs) of pixel grayscale values for images captured at $20$, $30$, $40$, $75$, $100$, and $200$~lux. The logarithmic x-axis highlights the differences in low-intensity regions. 
    (b) Corresponding RGB images with their average grayscale values indicated.}
    \label{fig:dark_vis}
    \vskip-4ex
\end{figure}

Fig.~\ref{fig:dark_vis} presents the grayscale distribution and example RGB images under different illumination levels. Within the selected low-light illumination range (under $75$~lux), the majority of pixels lie in the low-intensity region ($<20$) in the $0 \sim 255$ grayscale range, and objects are barely visible in corresponding images. 
Statistical analysis and visualization confirm that the selected illumination range constitutes low-light conditions under our imaging configuration.

\subsection{Dataset Collection}

All training data are collected on the SO100+DAVIS346 platform through teleoperation. During data collection, target object positions are randomized across the workspace with approximately balanced sampling to avoid spatial bias.
For Pick-Place, three object colors (red, yellow, and green) are used. For Sorting and Stacking, the positions of the red and green objects are swapped with equal probability to prevent fixed color–location associations.
To improve recovery from off-trajectory states, each task additionally includes $20$ episodes in which the robot arm starts from a random pose. These recovery episodes are recorded under normal illumination ($200$~lux).
In total, the dataset contains $724$ teleoperated episodes ($310$ Pick-Place, $244$ Sorting, and $170$ Stacking), comprising approximately $339k$ frames. 
Episodes last about $10$ seconds for Pick-Place and Stacking, and $15$ seconds for Sorting.

\subsection{Network and Training Details}
The event adapter adopts a ViT-style encoder similar to the image branch. Compared with the image encoder (SigLIP), the feature dimension is reduced from $768$ to $384$, and the number of layers is reduced from $12$ to $4$. For feature fusion, we also employ a lightweight design based on channel concatenation followed by an MLP with a hidden dimension of $1536$. Before concatenation, the event features are projected to $768$ to preserve identical dimensions with image features. The fused features are then fed into the next encoder layer.

\noindent \textbf{Image-based methods, E2VID and event overlay.} Enhanced images are generated offline and used in place of the original RGB input. For E2VID, the reconstructed image is used together with the RGB image as inputs. Then we freeze the VLM backbone and fine-tune only the action expert and projection layers to adapt the pretrained model to our deployment environment, denoted as \emph{Action} stage. Training runs for $20k$ iterations with a learning rate of $2 \times 10^{-4}$ and a batch size of $256$, while keeping other settings identical to the official implementation. 

\noindent \textbf{Event adapter.} 
For the event adapter approach, training proceeds in two additional stages. 
We first train the newly introduced event adapter and fusion module from scratch for $10k$ iterations (learning rate $5 \times 10^{-4}$, batch size $128$) while freezing the remaining components, denoted as \emph{Event} stage. We then jointly fine-tune the event adapter, fusion module, action expert, and projection layers for another $10k$ iterations (learning rate $1\times10^{-4}$, batch size $128$), denoted as \emph{Joint} stage.
For fair comparison, training schedules for all methods are chosen to preserve stable convergence; additional iterations do not further improve performance.

\section{Additional Experiments}
\label{Sec:addi_exp}

\subsection{Extremely Low Illumination Performance}

Tab.~\ref{tab:s1_detailed_illumination} reports the success rates of E-VLA under extremely low illumination below 20~lux. In this regime, RGB images are fully black-clipped, causing all image-based methods to fail (approximately $0\%$ success rate), and perception relies almost entirely on events. Nevertheless, E-VLA maintains a stable success rate of $50 \sim 80\%$ down to $4$~lux. Performance only degrades significantly at $2$~lux, where sensor noise begins to dominate the signal. 
These results indicate that event signals alone still provide useful cues for simple manipulation tasks such as Pick-Place, and offer substantially greater robustness and usability in extremely low-light environments compared to image-based approaches.

\begin{table}[h]
  \centering
  \footnotesize
  \setlength{\tabcolsep}{4pt}
  \renewcommand{\arraystretch}{1.15}
  \caption{Success rates (\%) under detailed lower illumination levels on the Pick-Place task. Higher is better. (Unlisted methods show a near-zero success rate.)}
  \label{tab:s1_detailed_illumination}
  \begin{tabular}{@{} l *{6}{c} S[table-format=2.1] @{}}
    \toprule
    \rowcolor{tblhead}
    \multicolumn{1}{l}{\textbf{Method}} &
    \multicolumn{1}{c}{\textbf{15~lux}} &
    \multicolumn{1}{c}{\textbf{10~lux}} &
    \multicolumn{1}{c}{\textbf{8~lux}}  &
    \multicolumn{1}{c}{\textbf{6~lux}}  &
    \multicolumn{1}{c}{\textbf{4~lux}}  &
    \multicolumn{1}{c}{\textbf{2~lux}}  &
    \multicolumn{1}{c}{\textbf{Average}} \\
    \midrule
    \midrule
    \multicolumn{1}{l}{Ours overlay}     & 65 & 60 & 65 & 60 & 50 & 30 & 55.0 \\
    \multicolumn{1}{l}{Ours adapter}     & 95 & 85 & 90 & 85 & 80 & 35 & 78.3 \\
    \bottomrule
  \end{tabular}%
  \vskip-2em
\end{table}

\subsection{Analysis for E2VID}

\begin{figure}[t]
    \centering
    \includegraphics[width=\linewidth]{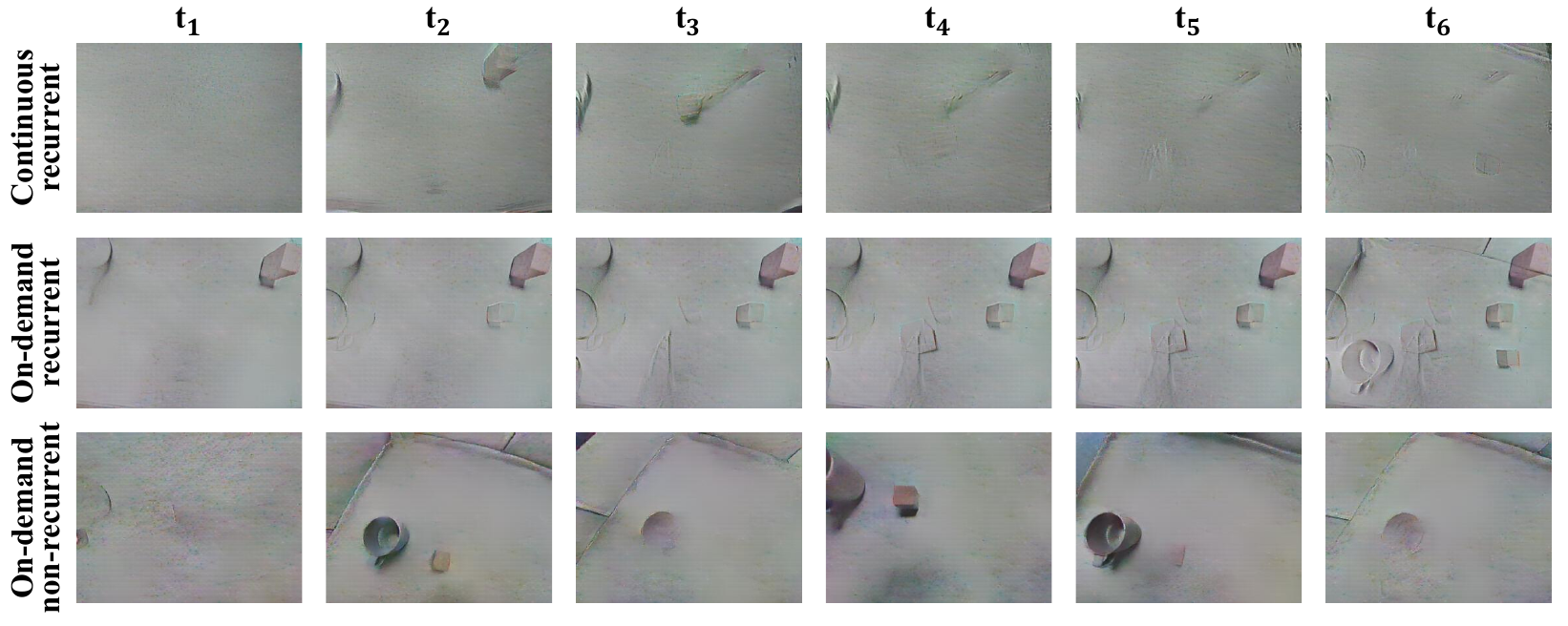}
    \vskip-2ex
    \caption{Reconstruction results of different E2VID implementations under $40$~lux. Each row shows a continuous temporal sequence ($\text{t}_1\sim\text{t}_6$) corresponding to six consecutive policy inference steps.}
    \label{fig:e2vid_vis}
    \vskip-4ex
\end{figure}

E2VID reconstructs images from event streams using a recurrent hidden state that is updated as new events arrive. 
To integrate this process into the E-VLA pipeline, a straightforward implementation is to run E2VID continuously so that the hidden state is always updated. However, this strategy requires high-frequency reconstruction, which significantly slows down the manipulation. An alternative design is to execute E2VID only when the policy requires a new observation, reducing reconstruction frequency and improving runtime efficiency. This on-demand setting leads to two variants: recurrent and non-recurrent.
In all implementations, the reconstructed image is fed into the VLA backbone as a second-view input, and the VLA model is pretrained in the same way. The success rates of the three implementations are reported in Tab.~\ref{tab:s2_e2vid_analysis}, with corresponding reconstruction results shown in Fig.~\ref{fig:e2vid_vis}.

% 这个表格好像画成图会更好

\begin{table}[h]
  \centering
  \footnotesize
  \setlength{\tabcolsep}{4pt}
  \renewcommand{\arraystretch}{1.15}
  \caption{Success rates (\%) for different implementations of E2VID. Results are reported under different illumination levels on the Pick-Place task. Higher is better.}
  \label{tab:s2_e2vid_analysis}
  \resizebox{\textwidth}{!}{%
  \begin{tabular}{@{} l *{6}{c} S[table-format=2.1] @{}}
    \toprule
    \rowcolor{tblhead}
    \multicolumn{1}{l}{\textbf{Implementation}} &
    \multicolumn{1}{c}{\textbf{75~lux}} &
    \multicolumn{1}{c}{\textbf{40~lux}} &
    \multicolumn{1}{c}{\textbf{35~lux}} &
    \multicolumn{1}{c}{\textbf{30~lux}} &
    \multicolumn{1}{c}{\textbf{25~lux}} &
    \multicolumn{1}{c}{\textbf{20~lux}} &
    \multicolumn{1}{c}{\textbf{Average}} \\ 
    \midrule
    \midrule
    \multicolumn{1}{l}{Continuous recurrent}      & 80 & 60 & 55 & 10 & 5 & 5 & 35.8 \\
    \multicolumn{1}{l}{On-demand recurrent}    & 80 & 55 & 50 & 40 & 15 & 10 & 41.7\\
    \multicolumn{1}{l}{On-demand non-recurrent}   & 100 & 85 & 60 & 40 & 25 & 10 & 53.3 \\
    \bottomrule
  \end{tabular}%
}
\end{table}

Continuous recurrent reconstruction produces low-contrast and blurry images due to the reduced event rate caused by slow system operation. 
The on-demand recurrent variant generates more distinct images, but preserving the recurrent state introduces duplication artifacts between successive inferences, which confuses the policy. Removing the recurrent connection eliminates this artifact and yields the clearest reconstructions among the three variants. However, the absence of temporal accumulation, together with the unstable event rate during manipulation, leads to discontinuities and intermittent signal loss. 
As a result, the three implementations achieve only $30 \sim 50\%$ success rates on average.

These observations suggest that reconstruction-based pipelines introduce additional computational overhead and implicitly assume temporally stable event streams. However, in VLA manipulation, event generation is tightly coupled with robot actions and inference frequency, resulting in irregular event dynamics that reduce the reliability of reconstruction-based event perception.

\subsection{Other VLA Backbones}
We replicate our experiments with the $\pi_{0.5}$ model. As shown in Table~\ref{tab:r2_pi}, performance degradation under low-light conditions persists and cannot be addressed by scaling model capacity alone. Our method consistently outperforms the image-only baseline. It suggests that our method can generalize across different architectures of VLA backbones.
\begin{table}[h]
  \centering
  \footnotesize
  \setlength{\tabcolsep}{4pt}
  \renewcommand{\arraystretch}{1.15}
  \caption{Task success rates (\%) on $\pi_{0.5}$ backbone. Results are reported under different illumination levels on the Pick-Place task, higher is better. (20 trails)}
  \label{tab:r2_pi}
  \resizebox{\linewidth}{!}{%
  \begin{tabular}{@{} l *{6}{c} S[table-format=2.1] @{}}
    \toprule
    \rowcolor{headergray}
    \multicolumn{1}{l}{\textbf{Method}} &
    \multicolumn{1}{c}{\textbf{75~lux}} &
    \multicolumn{1}{c}{\textbf{40~lux}} &
    \multicolumn{1}{c}{\textbf{35~lux}} &
    \multicolumn{1}{c}{\textbf{30~lux}} &
    \multicolumn{1}{c}{\textbf{25~lux}} &
    \multicolumn{1}{c}{\textbf{20~lux}} &
    \multicolumn{1}{c}{\textbf{Average}} \\
    \midrule
    \multicolumn{1}{l}{Image baseline}  & 100.0 & 87.5 & 85.0 & 52.5 & 10.0 & 0.0 & 55.8 \\
    \multicolumn{1}{l}{Event adapter}   & 100.0 & 100.0 & 95.0 & 92.5 & 92.5 & 90.0 & 95.0 \\
    \bottomrule
  \end{tabular}%
  }
\end{table}

\section{Efficiency Analysis}
\label{Sec:effi}

To analyze the computational overhead, we report the additional parameters and FLOPs introduced by each method compared to the image baseline.
As shown in Table~\ref{tab:s3_efficiency}, the proposed methods (Overlay and Adapter) introduce only a small computational overhead compared to enhancement and reconstruction methods. 
The simple overlay strategy directly fuses events with RGB inputs and therefore introduces no additional parameters and negligible computation. 
Our learnable adapter module is also lightweight, adding only $13.3M$ parameters and $20.4G$ FLOPs, which is substantially lower than several existing event-based approaches such as EvLight and E2VID. 
These results demonstrate that our methods achieve effective event utilization while maintaining favorable computational efficiency.

\begin{table}[h]
  \centering
  \footnotesize
  \setlength{\tabcolsep}{4pt}
  \renewcommand{\arraystretch}{1.15}
  \caption{Additional parameters and FLOPs compared to the image baseline. E2VID is measured on a single inference with 2000 events, while other methods are measured with a $260 \times 346$ input resolution. Lower is better. Abbreviations: R-Net (RetinexNet), R-former (Retinexformer).}
  \label{tab:s3_efficiency}
  \resizebox{\textwidth}{!}{%
  \begin{tabular}{@{} l *{7}{S[table-format=3.1, round-precision=1]} @{}}
    \toprule
    \rowcolor{tblhead}
    \multicolumn{1}{l}{\textbf{Metric}} &
    \multicolumn{1}{c}{\textbf{R-Net}} &
    \multicolumn{1}{c}{\textbf{R-former}}  &
    \multicolumn{1}{c}{\textbf{EvLight}}  &
    \multicolumn{1}{c}{\textbf{E2VID}}  &
    \multicolumn{1}{c}{\textbf{Overlay}}  &
    \multicolumn{1}{c}{\textbf{Adapter}} \\
    \midrule
    \midrule
    \multicolumn{1}{l}{Params (M) $\downarrow$}     
    & 0.56 & 1.61 & 22.73 & 10.71 & 0.00 & 13.31 \\
    \multicolumn{1}{l}{FLOPs (G) $\downarrow$}     
    & 48.5 & 23.5 & 533.7 & 171.5 & \sim 0 & 20.4 \\
    \bottomrule
  \end{tabular}%
}
  \vskip-2em
\end{table}

\section{Additional Ablations}
\label{Sec:ablation}

\subsection{Ablations on Training Augmentation}

\begin{table}[h]
  \centering
  \footnotesize
  \setlength{\tabcolsep}{4pt}
  \renewcommand{\arraystretch}{1.15}
  \caption{Task success rates (\%) on training augmentation. Results are reported under different illumination levels on the Pick-Place task, higher is better. (20 trails)}
  \label{tab:r1_augment}
  \resizebox{\linewidth}{!}{%
  \begin{tabular}{@{} l *{6}{c} S[table-format=2.1] @{}}
    \toprule
    \rowcolor{headergray}
    \multicolumn{1}{l}{\textbf{Method}} &
    \multicolumn{1}{c}{\textbf{75~lux}} &
    \multicolumn{1}{c}{\textbf{40~lux}} &
    \multicolumn{1}{c}{\textbf{35~lux}} &
    \multicolumn{1}{c}{\textbf{30~lux}} &
    \multicolumn{1}{c}{\textbf{25~lux}} &
    \multicolumn{1}{c}{\textbf{20~lux}} &
    \multicolumn{1}{c}{\textbf{Average}} \\
    \midrule
    \multicolumn{1}{l}{Image baseline}      & 100.0 & 77.5 & 72.5 & 35.0 & 2.5 & 0.0 & 47.9 \\
    \multicolumn{1}{l}{Image augmentation}  & 100.0 & 82.5 & 75.0 & 40.0 & 12.5 & 0.0 & 51.7 \\
    \multicolumn{1}{l}{Event adapter}       & 100.0 & 97.5 & 95.0 & 90.0 & 90.0 & 87.5 & 93.3 \\
    \bottomrule
  \end{tabular}%
  }
\end{table}

We apply aggressive data augmentations (random gamma brightness reduction (1.1-2.8) and random directional blurring (3-15 pixels)) on 80\% of all training data. Tab.~\ref{tab:r1_augment} shows that augmentations only provide a slight benefit compared to the image baseline. It suggests that software augmentations cannot fully handle information loss, supporting the necessity of introducing the event modality under degraded conditions.

\subsection{Ablations on Event Representation}

\begin{table}[h]
  \centering
  \footnotesize
  \setlength{\tabcolsep}{4pt}
  \renewcommand{\arraystretch}{1.12}
  \caption{Task success rates (\%) for various event representation. Results are reported under different illumination levels (lux) on the Pick-Place task. Higher is better.}
  \label{tab:s4_event_repr}
  \begin{tabular}{@{} l *{6}{c} S[table-format=2.1] }
    \toprule
    \rowcolor{tblhead}
    \multicolumn{1}{l}{\textbf{Event Repr.}} &
    \multicolumn{1}{c}{\textbf{75~lux}} &
    \multicolumn{1}{c}{\textbf{40~lux}} &
    \multicolumn{1}{c}{\textbf{35~lux}} &
    \multicolumn{1}{c}{\textbf{30~lux}} &
    \multicolumn{1}{c}{\textbf{25~lux}} &
    \multicolumn{1}{c}{\textbf{20~lux}} &
    \multicolumn{1}{c}{\textbf{Average}}\\ 
    \midrule
    \midrule
    \multicolumn{1}{l}{Voxel grid~\cite{zihao2018unsupervised}}      & 90 & 80 & 85 & 80 & 60 & 35 & 71.7 \\
    \multicolumn{1}{l}{Time surface~\cite{lagorce2016timesurface}}    & 95 & 85 & 80 & 80 & 75 & 80 & 82.5 \\
    \multicolumn{1}{l}{Accu. Sum}       & 90 & 85 & 80 & 80 & 75 & 65 & 79.2 \\
    \rowcolor{oursrow}
    \multicolumn{1}{l}{Accu. Count}     & 100 & 100 & 95 & 90 & 90 & 90 & 94.2\\
    \bottomrule
  \end{tabular}
\end{table}

Tab.~\ref{tab:s4_event_repr} compares three common event representation schemes: voxel grid~\cite{zihao2018unsupervised}, time surface~\cite{lagorce2016timesurface}, and event accumulation (polarity-aware sum and polarity-agnostic count). We observe that the count-based accumulation tends to perform better than the sum variant, likely because it produces clearer and more stable edge structures by avoiding polarity cancellation. In addition, accumulated event representations generally achieve better results than the other approaches, suggesting that frame-like representations could better leverage the pretrained knowledge in the VLM.

\subsection{Ablation on Training Regularization}

\begin{figure}[h]
    \centering
    \includegraphics[width=0.5\linewidth]{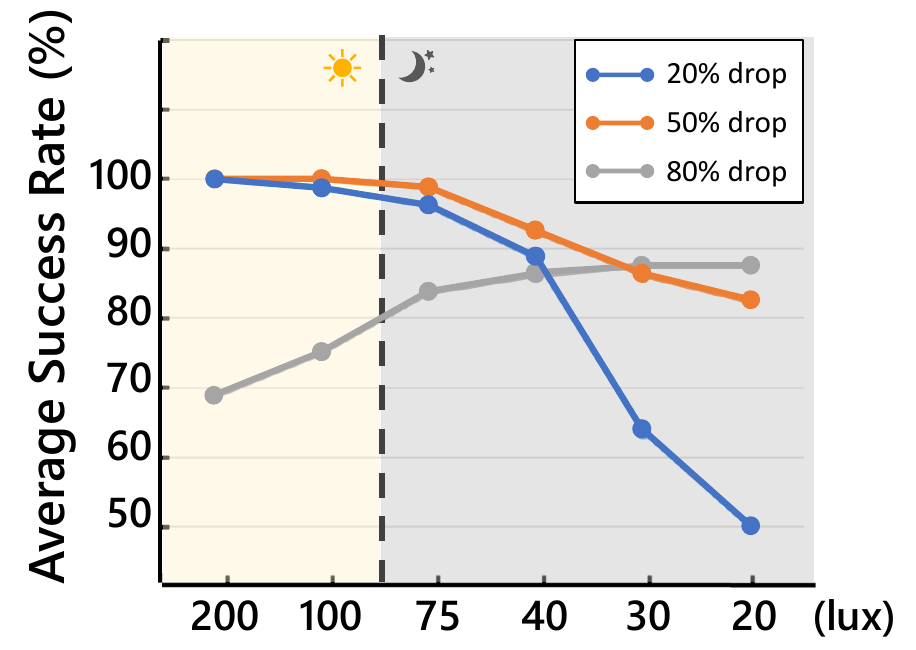}
    \caption{
    Average success rates of Pick-Place and Sorting, using event adapter method training on different image dropout rates ($20 \sim 80\%$). Results are reported under different illumination levels (left: normal, right: low-light). Higher is better.}
    \label{fig:dropout_rate}
    \vskip-4ex
\end{figure}

To prevent shortcuts on image cues during event adapter training, we introduce a stochastic image dropout strategy that randomly blocks image inputs with a given probability during training. As shown in Fig.~\ref{fig:dropout_rate}, low dropout rates ($20\%$) result in limited event utilization and marginal gains under low-light conditions. Increasing the block rate to $50\%$ strengthens reliance on event features and improves robustness in the low-illumination range. 
However, an excessively high dropout rate ($80\%$) degrades performance under normal and slightly low lighting, which can be largely attributed to the failure to utilize bright images and distinguish colors in the Sorting task. This result suggests that aggressive dropout can hinder the retention and use of pretrained visual knowledge. We therefore adopt a dropout rate of $50\%$ as a practical trade-off between low-light robustness and well-lit performance.
\vskip-3ex

\section{Additional Visualization}
\label{Sec:mani_seq}

\begin{figure}[t]
    \centering
    \includegraphics[width=\linewidth]{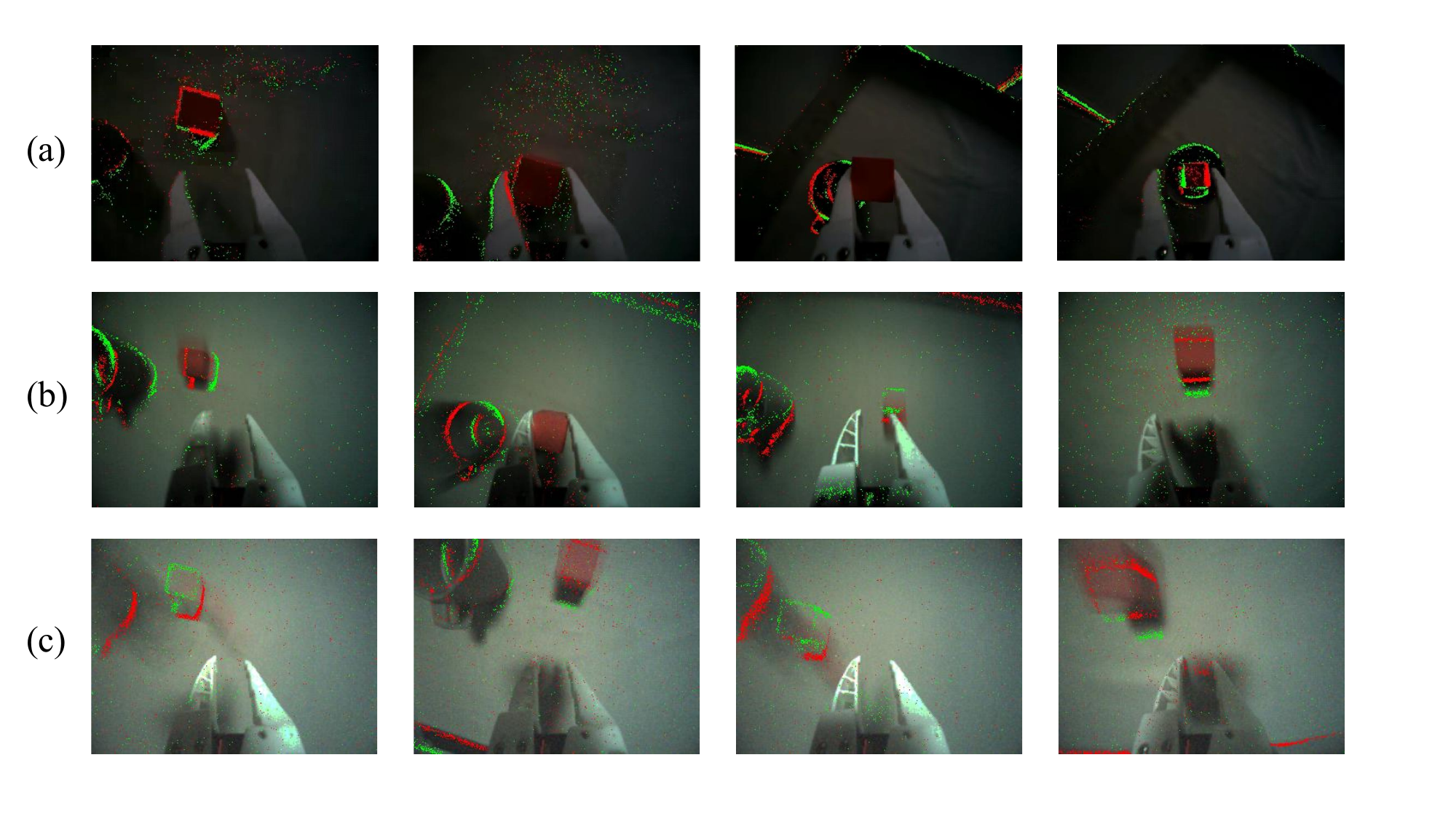}
    \vskip-4ex
    \caption{(a): Pick-Place example sequence. (b) and (c): 300ms and 1000ms motion blur scenes.}
    \label{fig:visual}
\end{figure}

As shown in Fig.~\ref{fig:visual}, under low-light conditions, events provide clear edge cues that facilitate task execution. In the presence of severe motion blur, image edges become indistinct, whereas events are inherently insensitive to motion blur and can therefore offer reliable spatial guidance.

\section{Limitations and Potential Solutions}
\label{Sec:limit_solution}

Although E-VLA performs well in many low-light and motion-blurred scenarios, several limitations remain. First, event cameras capture intensity changes caused by brightness variation or edge motion and are therefore insensitive to semantic cues with uniform intensity, such as color. Second, the monocular wrist-mounted camera can be partially occluded by manipulated objects, which restricts the observable region. These factors lead to color misclassification in the Sorting task under low illumination and reduced precision in the Stacking task, as shown in Tab.~\ref{tab:s5_sorting_stacking}. 
A detailed analysis is provided in Sec.~5.7 of the main paper. 
In addition, the current E-VLA mainly responds to motion events induced by robot manipulation, making scene perception and utilization largely passive. Finally, the generalization ability to more complex scenes remains unclear due to the scarcity and diversity of event-based training data. 

\begin{table}[h]
  \centering
  \footnotesize
  \setlength{\tabcolsep}{2pt}
  \renewcommand{\arraystretch}{1.25}
  \caption{Task success rates (\%) on the Sorting and Stacking task. Results are reported under different illumination levels. Higher is better. }
  \label{tab:s5_sorting_stacking}
  \resizebox{\textwidth}{!}{%
    \begin{tabular}{@{} l *{12}{S[table-format=3.1, round-mode=places, round-precision=1]}}
      \toprule
      \rowcolor{tblhead}
      \multicolumn{1}{c}{} &
      \multicolumn{2}{c}{\textbf{75~lux}} &
      \multicolumn{2}{c}{\textbf{40~lux}} &
      \multicolumn{2}{c}{\textbf{35~lux}} &
      \multicolumn{2}{c}{\textbf{30~lux}} &
      \multicolumn{2}{c}{\textbf{25~lux}} &
      \multicolumn{2}{c}{\textbf{20~lux}} \\
      \rowcolor{tblhead}
      \multicolumn{1}{c}{\textbf{Method}} &
      \multicolumn{1}{c}{Sorting} & \multicolumn{1}{c}{Stacking} &
      \multicolumn{1}{c}{Sorting} & \multicolumn{1}{c}{Stacking} &
      \multicolumn{1}{c}{Sorting} & \multicolumn{1}{c}{Stacking} &
      \multicolumn{1}{c}{Sorting} & \multicolumn{1}{c}{Stacking} &
      \multicolumn{1}{c}{Sorting} & \multicolumn{1}{c}{Stacking} &
      \multicolumn{1}{c}{Sorting} & \multicolumn{1}{c}{Stacking} \\
      \cmidrule(l){2-3} \cmidrule(l){4-5} \cmidrule(l){6-7}
      \cmidrule(l){8-9} \cmidrule(l){10-11} \cmidrule(l){12-13}
      \multicolumn{1}{l}{Image}
      & 100.0 & 60 & 85.0 & 45 & 62.5 & 35 & 30.0 & 15 & 0.0 & 0 & 0.0 & 0 \\
      \rowcolor{oursrow}
      \multicolumn{1}{l}{Ours overlay}
      & 95 & 55 & 80 & 60 & 72.5 & 45 & 70 & 40 & 67.5 & 35 & 50 & 30 \\
      \rowcolor{oursrow}
      \multicolumn{1}{l}{Ours adapter}
      & 97.5 & 60 & 85 & 55 & 82.5 & 50 & 82.5 & 55 & 80.0 & 45 & 70.0 & 40 \\
      \bottomrule
    \end{tabular}%
  }
\end{table}

Several potential solutions may help alleviate these issues. Prior studies suggest that specially designed demosaicing methods~\cite{color4e} can improve color reconstruction and provide richer chromatic information. Then, the blind spots of the wrist-mounted camera could be mitigated through improved hardware placement, such as more suitable mounting positions or mirrors near the gripper. Integrating actively modulated perception with a third-person viewpoint could further expand the observable workspace and provide richer scene representations, including depth maps or temporal-mapping images~\cite{bao2024temporal}. 
Finally, generalization to complex environments may benefit from larger and more diverse datasets, as well as transfer learning or other adaptation strategies.

\end{document}